%% file: main.tex
\PassOptionsToPackage{table}{xcolor}
\documentclass[10pt,twocolumn,letterpaper]{article}
\usepackage[accsupp]{axessibility}  
\usepackage{cvpr}              
\usepackage{multirow}
\usepackage[normalem]{ulem}
\usepackage{bm}
\usepackage{tcolorbox}
\usepackage{footmisc}
\PassOptionsToPackage{table}{xcolor}
\usepackage{multirow}
\usepackage{cvpr}

\usepackage{graphicx}
\usepackage{amsmath}
\usepackage{amssymb}
\usepackage{booktabs}
\usepackage{bbding}

\newcommand\blfootnote[1]{%
  \begingroup
  \renewcommand\thefootnote{}\footnote{#1}%
  \addtocounter{footnote}{-1}%
  \endgroup
}
\definecolor{cvprblue}{rgb}{0.21,0.49,0.74}
\usepackage[pagebackref,breaklinks,colorlinks,allcolors=cvprblue]{hyperref}
\usepackage[dvipsnames]{xcolor}

\def\paperID{23307} 
\def\confName{CVPR}
\def\confYear{2026}

\title{OpenVO: Open-World Visual Odometry with Temporal Dynamics Awareness}

\author{
Phuc Nguyen$^{*}$ \quad Anh N. Nhu$^{*}$ \quad Ming C. Lin \\
\normalsize{University of Maryland, College Park, USA} \\
\tt\small \{phucnda, anhu, lin\}@umd.edu  \\
{\url{https://openvo.github.io/}}
}

\def\Approach{OpenVO}

\begin{document}
\input{definition}

\twocolumn[{
\renewcommand\twocolumn[1][]{#1}%
\maketitle
\vspace{-20pt}
\begin{center}%
\includegraphics[width=.80\linewidth]{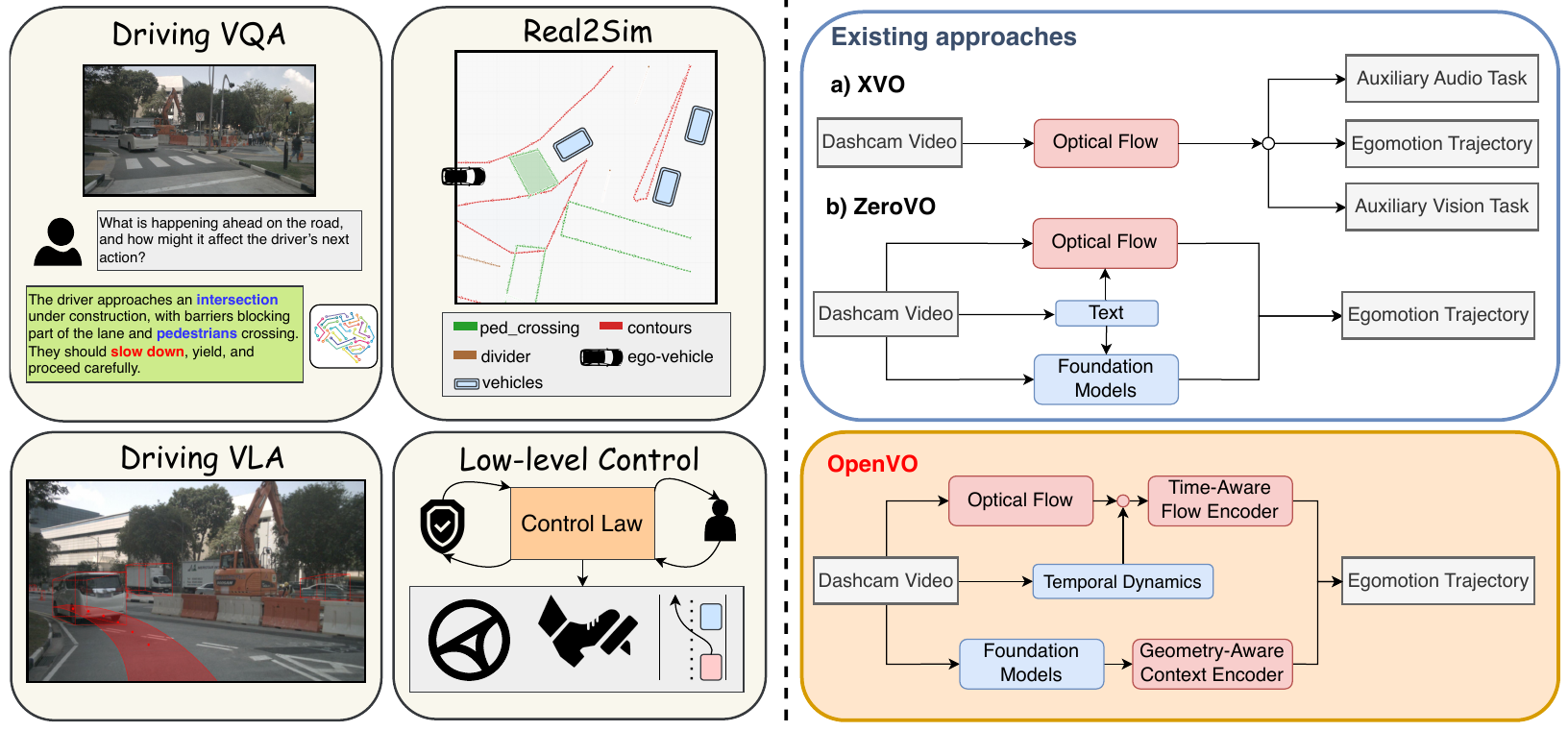}
\vspace{-8pt}
\captionof{figure}{\textbf{Left:} Generalized Visual Odometry provides real-world ego-motion and trajectory estimates that bridge perception and control in autonomous driving. It enables scene understanding (Driving VQA \cite{qian2024nuscenesqa, wei2025driveqa}), simulation (Real2Sim \cite{liu2023vectormapnet, shi2024globalmapnet}), action grounding (Driving VLA \cite{hwang2024emma, li2025drivevla}), and precise motion feedback for low-level control \cite{kim2020advisable, chen2022learning, xu2025drivegpt4}. \textbf{Right:} We introduce \textbf{\Approach,} a generalizable visual odometry framework that estimates real-world ego-motion from uncalibrated dashcam videos and remains robust across varying observation rates. Our design integrates our ``Time-Aware Flow Encoder'' for modeling temporal dynamics and a ``Geometry-Aware Context Encoder'' for extracting consistent scene geometry, enabling robust and generalizable motion estimation across diverse visual and temporal domains. 
}
\label{fig:teaser}
\end{center}
}]

\input{sec/0_abstract}    
\input{sec/1_intro}
\input{sec/2_relatedwork}

\input{sec/3_method}
\input{sec/4_experiments}

\input{sec/5_conclusion} 
\input{supp_mat}

    \bibliographystyle{ieeenat_fullname}
    \bibliography{main}


\end{document}

%% file: definition.tex
\def\mA{\mathcal{A}}
\def\mB{\mathcal{B}}
\def\mC{\mathcal{C}}
\def\mD{\mathcal{D}}
\def\mE{\mathcal{E}}
\def\mF{\mathcal{F}}
\def\mG{\mathcal{G}}
\def\mH{\mathcal{H}}
\def\mI{\mathcal{I}}
\def\mJ{\mathcal{J}}
\def\mK{\mathcal{K}}
\def\mL{\mathcal{L}}
\def\mM{\mathcal{M}}
\def\mN{\mathcal{N}}
\def\mO{\mathcal{O}}
\def\mP{\mathcal{P}}
\def\mQ{\mathcal{Q}}
\def\mR{\mathcal{R}}
\def\mS{\mathcal{S}}
\def\mT{\mathcal{T}}
\def\mU{\mathcal{U}}
\def\mV{\mathcal{V}}
\def\mW{\mathcal{W}}
\def\mX{\mathcal{X}}
\def\mY{\mathcal{Y}}
\def\mZ{\mathcal{Z}} 

\def\bbN{\mathbb{N}} 
\def\bbR{\mathbb{R}} 
\def\bbP{\mathbb{P}} 
\def\bbQ{\mathbb{Q}} 
\def\bbE{\mathbb{E}}

\def\1n{\mathbf{1}_n}
\def\0{\mathbf{0}}
\def\1{\mathbf{1}}

\def\A{{\bf A}}
\def\B{{\bf B}}
\def\C{{\bf C}}
\def\D{{\bf D}}
\def\E{{\bf E}}
\def\F{{\bf F}}
\def\G{{\bf G}}
\def\H{{\bf H}}
\def\I{{\bf I}}
\def\J{{\bf J}}
\def\K{{\bf K}}
\def\L{{\bf L}}
\def\M{{\bf M}}
\def\N{{\bf N}}
\def\O{{\bf O}}
\def\P{{\bf P}}
\def\Q{{\bf Q}}
\def\R{{\bf R}}
\def\S{{\bf S}}
\def\T{{\bf T}}
\def\U{{\bf U}}
\def\V{{\bf V}}
\def\W{{\bf W}}
\def\X{{\bf X}}
\def\Y{{\bf Y}}
\def\Z{{\bf Z}}

\def\a{{\bf a}}
\def\b{{\bf b}}
\def\c{{\bf c}}
\def\d{{\bf d}}
\def\e{{\bf e}}
\def\f{{\bf f}}
\def\g{{\bf g}}
\def\h{{\bf h}}
\def\i{{\bf i}}
\def\j{{\bf j}}
\def\k{{\bf k}}
\def\l{{\bf l}}
\def\m{{\bf m}}
\def\n{{\bf n}}
\def\o{{\bf o}}
\def\p{{\bf p}}
\def\q{{\bf q}}
\def\r{{\bf r}}
\def\s{{\bf s}}
\def\t{{\bf t}}
\def\u{{\bf u}}
\def\v{{\bf v}}
\def\w{{\bf w}}
\def\x{{\bf x}}
\def\y{{\bf y}}
\def\z{{\bf z}}

\def\balpha{\mbox{\boldmath{$\alpha$}}}
\def\bbeta{\mbox{\boldmath{$\beta$}}}
\def\bdelta{\mbox{\boldmath{$\delta$}}}
\def\bgamma{\mbox{\boldmath{$\gamma$}}}
\def\blambda{\mbox{\boldmath{$\lambda$}}}
\def\bsigma{\mbox{\boldmath{$\sigma$}}}
\def\btheta{\mbox{\boldmath{$\theta$}}}
\def\bomega{\mbox{\boldmath{$\omega$}}}
\def\bxi{\mbox{\boldmath{$\xi$}}}
\def\bnu{\mbox{\boldmath{$\nu$}}}                                  
\def\bphi{\mbox{\boldmath{$\phi$}}}
\def\bmu{\mbox{\boldmath{$\mu$}}}

\def\bDelta{\mbox{\boldmath{$\Delta$}}}
\def\bOmega{\mbox{\boldmath{$\Omega$}}}
\def\bPhi{\mbox{\boldmath{$\Phi$}}}
\def\bLambda{\mbox{\boldmath{$\Lambda$}}}
\def\bSigma{\mbox{\boldmath{$\Sigma$}}}
\def\bGamma{\mbox{\boldmath{$\Gamma$}}}
                                  
\newcommand{\myprob}[1]{\mathop{\mathbb{P}}_{#1}}

\newcommand{\myexp}[1]{\mathop{\mathbb{E}}_{#1}}

\newcommand{\mydelta}[1]{1_{#1}}

\newcommand{\myminimum}[1]{\mathop{\textrm{minimum}}_{#1}}
\newcommand{\mymaximum}[1]{\mathop{\textrm{maximum}}_{#1}}    
\newcommand{\mymin}[1]{\mathop{\textrm{minimize}}_{#1}}
\newcommand{\mymax}[1]{\mathop{\textrm{maximize}}_{#1}}
\newcommand{\mymins}[1]{\mathop{\textrm{min.}}_{#1}}
\newcommand{\mymaxs}[1]{\mathop{\textrm{max.}}_{#1}}  
\newcommand{\myargmin}[1]{\mathop{\textrm{argmin}}_{#1}} 
\newcommand{\myargmax}[1]{\mathop{\textrm{argmax}}_{#1}} 
\newcommand{\myst}{\textrm{s.t. }}

\newcommand{\denselist}{\itemsep -1pt}
\newcommand{\sparselist}{\itemsep 1pt}

\definecolor{pink}{rgb}{0.9,0.5,0.5}
\definecolor{purple}{rgb}{0.5, 0.4, 0.8}   
\definecolor{gray}{rgb}{0.3, 0.3, 0.3}
\definecolor{mygreen}{rgb}{0.2, 0.6, 0.2}

\newcommand{\cyan}[1]{\textcolor{cyan}{#1}}
\newcommand{\blue}[1]{\textcolor{blue}{#1}}
\newcommand{\magenta}[1]{\textcolor{magenta}{#1}}
\newcommand{\pink}[1]{\textcolor{pink}{#1}}
\newcommand{\green}[1]{\textcolor{green}{#1}} 
\newcommand{\gray}[1]{\textcolor{gray}{#1}}    
\newcommand{\mygreen}[1]{\textcolor{mygreen}{#1}}    
\newcommand{\purple}[1]{\textcolor{purple}{#1}}       

\definecolor{greena}{rgb}{0.4, 0.5, 0.1}
\newcommand{\greena}[1]{\textcolor{greena}{#1}}

\definecolor{bluea}{rgb}{0, 0.4, 0.6}
\newcommand{\bluea}[1]{\textcolor{bluea}{#1}}
\definecolor{reda}{rgb}{0.6, 0.2, 0.1}
\newcommand{\reda}[1]{\textcolor{reda}{#1}}

\def\changemargin#1#2{\list{}{\rightmargin#2\leftmargin#1}\item[]}
\let\endchangemargin=\endlist
                                               
\newcommand{\cm}[1]{}

\newcommand{\mhoai}[1]{{\color{magenta}\textbf{[MH: #1]}}}

\newcommand{\mtodo}[1]{{\color{red}$\blacksquare$\textbf{[TODO: #1]}}}
\newcommand{\myheading}[1]{\vspace{1ex}\noindent \textbf{#1}}
\newcommand{\htimesw}[2]{\mbox{$#1$$\times$$#2$}}


\newif\ifshowsolution
\showsolutiontrue

\ifshowsolution  
\newcommand{\Solution}[2]{\paragraph{\bf $\bigstar $ SOLUTION:} {\sf #2} }
\newcommand{\Mistake}[2]{\paragraph{\bf $\blacksquare$ COMMON MISTAKE #1:} {\sf #2} \bigskip}
\else
\newcommand{\Solution}[2]{\vspace{#1}}
\fi

\newcommand{\truefalse}{
\begin{enumerate}
	\item True
	\item False
\end{enumerate}
}

\newcommand{\yesno}{
\begin{enumerate}
	\item Yes
	\item No
\end{enumerate}
}

\newcommand{\Sref}[1]{Sec.~\ref{#1}}
\newcommand{\Eref}[1]{Eq.~(\ref{#1})}
\newcommand{\Fref}[1]{Fig.~\ref{#1}}
\newcommand{\Tref}[1]{Table~\ref{#1}}

\definecolor{gray}{rgb}{0.3, 0.3, 0.3}

%% file: sec/0_abstract.tex
\begin{abstract}
We introduce \Approach, a novel framework for Open-world Visual Odometry (VO) with temporal awareness under limited input conditions. \Approach~effectively estimates real-world–scale ego-motion from monocular dashcam footage with varying observation rates and uncalibrated cameras, enabling robust trajectory dataset construction from rare driving events recorded in dashcam.
Existing VO methods are trained on fixed observation frequency (e.g., 10Hz or 12Hz), completely overlooking temporal dynamics information. Many prior methods also require calibrated cameras with known intrinsic parameters. 
Consequently, their performance degrades when (1) deployed under unseen observation frequencies or (2) applied to uncalibrated cameras. These significantly limit their generalizability to many downstream tasks, such as extracting trajectories from dashcam footage.
To address these challenges, \Approach~(1) explicitly encodes temporal dynamics information within a two-frame pose regression framework and (2) leverages 3D geometric priors derived from foundation models. We validate our method on three major autonomous-driving benchmarks -- KITTI, nuScenes, and Argoverse 2 -- achieving {\bf more than 20\%} performance improvement over state-of-the-art approaches. 
Under varying observation rate settings, our method is significantly more robust, achieving {\bf 46\%–92\%} lower errors across all metrics.
These results demonstrate the versatility of \Approach~ for real-world 3D reconstruction and diverse downstream applications.
\end{abstract}
\vspace{-14pt}


%% file: sec/1_intro.tex
\vspace*{-0.5em}
\section{Introduction} 
\vspace*{-0.5em}

Visual Odometry (VO) \cite{1315094} \blfootnote{$^*$: Equal contribution}system is an fundamental component in autonomous driving and robotics, providing crucial information about an agents' pose and ego-motion in world coordinates to enable comprehensive situational awareness and understanding. In many downstream tasks, both online (e.g. onboard perception systems \cite{maimone2007two, 9561714, 4651147}) and offline (e.g. trajectory log analysis and driving dataset construction \cite{9197053}), VO must be robust to various sources of variations, including but not limited to observation sampling rates and sensor calibrations, which present significant challenges to VO models. One particularly challenging case is trajectory reconstruction from dashcam footage obtained from online platforms such as YouTube. These videos contain rich spatial and dynamical information about rare, long-tail driving events (e.g., crashes) that are otherwise difficult to collect \textit{at scale} due to economic, legal, and ethical constraints.
 Dashcam videos are typically monocular and uncalibrated, exhibiting large variations in camera calibrations and poses across sources. Moreover, they could be recorded at different frame rates, introducing temporal inconsistencies that degrade model accuracy. The temporal dynamics implicitly encoded in the observation rate could be exploited to enhance VO accuracy and robustness. Prior works in VO largely overlooked this factor. Studies in sequential decision-making~\cite{thodoroff2022benchmarking, nhu2025timeaware} have demonstrated that such temporal inconsistencies can significantly degrade model performance when deployed on mismatched observation rates (e.g., trained on 20 Hz, yet deployed on 12 Hz). 

Classical geometry-based methods \cite{mur2015orb, mur2017orb, campos2021orb, forster2014svo, wang2017stereo} typically require access to accurate camera calibration at inference and struggle with uncalibrated observations. Early learning-based approaches~\cite{7347378, wang2017deepvo, li2018undeepvo, loo2019cnn, yang2020d3vo} are trained and evaluated on the same datasets under generally consistent sensor calibrations and without explicit awareness of camera geometry, limiting their generalizability to substantially different sensor setups. More recent works explicitly incorporate camera geometric priors into their framework \cite{wang2021tartanvo, teed2021droid, teed2023deep} to achieve strong VO estimation accuracy but still assume known camera intrinsics. The latest VO frameworks are more generalized via multi-modal self-training for intrinsic-free estimation \cite{lai2023xvo} or geometry-aware architectures with language priors for zero-shot generalization \cite{lai2025zerovo}. However, all existing methods are trained and evaluated on fixed frame rate, overlooking temporal dynamics from varying observation frequencies.

Motivated by these gaps, we aim to address an important question: \textbf{\textit{How can a VO system generalize to uncalibrated observations under varying observation rates with unknown camera calibrations in open-world scenarios?}} To tackle this question, we present \textbf{OpenVO}, a generalized visual odometry framework that elegantly leverages \textit{temporal dynamic information} and \textit{3D geometry priors} derived from inferred camera parameters to robustly estimate ego-motion from \textit{uncalibrated monocular observations}. The core contributions of OpenVO are as follows:
\begin{enumerate}
    \item \textbf{Temporal frequency incorporation:} \Approach~encodes video frame-rate information into a time-aware embedding that conditions the optical flow features to adapt to varying temporal dynamics. This explicit modeling of temporal frequency enriches motion representation, enabling accurate and robust ego-motion estimation across diverse frame rates and video sources.
    \item \textbf{Differentiable 2D-Guided 3D flow estimation:} we construct differentiable 3D flow fields from estimated 2D flow model and metric depth. These 3D flow fields are then fused with time-aware 2D flow features, further enhancing the accuracy of ego-motion estimation in world coordinates.
    \item \textbf{Geometry context awareness:} we explicitly leverage inferred camera intrinsics and homogeneous projective geometry to provide geometric awareness of the scene, enabling OpenVO to generalize effectively to uncalibrated observations with diverse camera configurations. 
\end{enumerate}

Our framework consistently achieves state-of-the-art visual odometry performance on various large-scale benchmarks, including nuScenes \cite{caesar2020nuscenes}, KITTI \cite{geiger2013vision}, and Argoverse-2 \cite{wilson2023argoverse2}, outperforming existing state-of-the-art methods by over 20\% global absolute trajectory error. Most notably, while prior methods exhibit substantial performance degradation when tested on unseen observation rates, OpenVO demonstrates significantly greater robustness against temporal dynamic variations. OpenVO demonstrates strong generalization to uncalibrated and out-of-domain observations across varying frame rates, an essential capability for extracting and analyzing high-quality trajectory data from dashcam footage. Because accurate ego-motion estimation is fundamental to robotics and autonomous driving, OpenVO can serve as an important component in a wide range of downstream tasks, including online vectorized mapping, trajectory reconstruction, 3D scene understanding, and motion forecasting.

%% file: sec/2_relatedwork.tex
\vspace*{-0.5em}
\section{Related Work}
\vspace*{-0.5em}
OpenVO leverages projective geometry in traditional visual odometry and geometric priors, combined with multi-time-scale learning insights, for generalizable visual odometry.

\myheading{Monocular Visual Odometry.} There are two primary VO approaches: (1) traditional geometry-based and (2) learning-based methods. Traditional geometry-based approaches~\cite{1315094, realtimevo}, such as MonoSLAM~\cite{davison2007monoslam}, ORB-SLAM~\cite{mur2015orb, mur2017orb, campos2021orb}, SVO~\cite{forster2014svo}, DSO~\cite{engel2017direct, wang2017stereo} employ projective geometry and feature matching pipelines for precise and interpretable motion estimation under calibrated observation settings. These methods heavily depend on known camera parameters, careful initialization, and engineered feature quality or photometric assumptions, which limit their generalization to uncalibrated, texture-poor, or highly dynamic real-world scenarios. To address such limitations, end-to-end learning-based VO methods directly learn the mapping between RGB observations and frame-to-frame motion~\cite{7347378, wang2017deepvo, li2018undeepvo, clark2017vinet, wang2018end}. Some hybrid approaches incorporate learned geometric priors into classical pipelines (e.g. using monocular depth modules for constraint or joint optimization), combining geometric priors with the flexibility of deep learning~\cite{loo2019cnn, yang2020d3vo, teed2023deep, zhou2017unsupervised, teed2021droid, yin2018geonet}. However, these works train and evaluate their models under similar conditions (e.g., same dataset with fixed camera settings and limited scenario diversity), which restricts their generalizability to in-the-wild scenarios and uncalibrated observations with notable domain gaps (e.g. camera settings and scenarios).

\myheading{Generalized Visual Odometry.} TartanVO~\cite{wang2021tartanvo} incorporates an intrinsic layer and up-to-scale loss to handle scale ambiguity and varying camera configurations, achieving strong cross-dataset performance. However, TartanVO relies on ground-truth intrinsics, limiting its applicability to uncalibrated observations. DINO-VO~\cite{azhari2025dino} leverages the pretrained DINOv2~\cite{oquab2024dinov2} vision foundation model for more robust and generalizable correspondence matching in challenging scenarios. XVO~\cite{lai2023xvo} advances calibration-free generalization through cross-modal self-training, leveraging large-scale pseudo-labeled uncalibrated YouTube dashcam videos to improve robustness. ZeroVO~\cite{lai2025zerovo} further introduces language and 3D geometric priors via estimated 3D flows for zero-shot generalization. Although XVO and ZeroVO achieve strong performance on unseen domains with varying sensor settings and camera parameters, they overlook the impact of frame rates, leading to degraded performance under unseen observation frequencies. In contrast, our OpenVO achieves state-of-the-art performance across datasets with diverse camera configurations and frame rates, without ground-truth camera intrinsics.

\myheading{Geometric Prior Models.} Geometric priors such as metric depth and camera intrinsics are crucial for 3D reconstruction and understanding \cite{ nguyen2024open3dis, nguyen2025open, nguyen2025any3dis} and improving scale consistency in visual odometry~\cite{lai2025zerovo, yang2020d3vo, teed2021droid}. 
%
%
Monocular foundation metric depth models~\cite{yin2023metric3d, hu2024metric3dv2, bochkovskiydepth, piccinelli2024unidepth, piccinelli2025unidepthv2, piccinelli2025unidepthv2, yang2024depth, yang2024depth2, bhat2023zoedepth} enable absolute-scale depth prediction with strong zero-shot generalization across divese domains through large-scale training. 
Among these, some rely on known camera parameters~\cite{yin2023metric3d, hu2024metric3dv2}, while others remove this assumption and even achieve competitive performance in estimating focal lengths from monocular observation~\cite{bochkovskiydepth}. However, ground-truth camera parameters are not always available in downstream tasks. To bridge this gap, \cite{zhu2023wildcamera, jin2023perspective, lee2021ctrl} have proposed dedicated models to estimate camera intrinsics from monocular observations. In OpenVO, we employ the WildCamera~\cite{zhu2023wildcamera} and Metric3Dv2~\cite{hu2024metric3dv2} to infer camera intrinsics and metric depth for 3D flow construction, guided by the 2D optical flow from MaskFlowNet~\cite{Zhao_2020_maskflownet}.


\myheading{Multi-Time-Scale Adaptive Learning.} Despite being one of the key quantities in temporally-driven problems, including video understanding, visual odometry, sequential decision making, and dynamical systems, observation rate (i.e., the time step size $\Delta t = 1/f$) is typically fixed during the training process and \textit{not modeled explicitly} by prior VO methods. Specifically, in frame-to-frame visual odometry, existing state-of-the-art models~\cite{lai2025zerovo, lai2023xvo, wang2021tartanvo} take a pair of consecutive frames $(I_{t-1}, I_{t})$, optionally along with auxiliary conditioning modality, such as language prior $c_{text}$, to estimate the ego motion. These models completely overlook the impacts of the temporal interval $\Delta t$ on the learning process, which might encode important temporal dynamics in visual odometry, such as velocity.  As a result, they are trained under a single time scale (e.g., a fixed sampling rate of 20 Hz), leading to a subtle form of \textit{temporal overfitting}~\cite{nhu2025timeaware} -- where the learned motion representation is implicitly optimized for that specific sampling rate -- and experiencing potential performance degradation when deployed under mismatched observation frequency (e.g., 12 Hz). Although these phenomena are underexplored in visual odometry, prior studies in object detection \cite{nguyen2025ha, zhang2020dynamic}, reinforcement learning~\cite{thodoroff2022benchmarking} and world models~\cite{nhu2025timeaware} have shown that temporal overfitting to a fixed observation rate significantly degrades model robustness under unseen time scales. Motivated by these findings, we introduce a novel time-aware flow encoder and train the model across a diverse range of observation frequencies, exposing it to multiple temporal dynamic scales. This strategy effectively improves OpenVO's accuracy across unseen large-scale autonomous driving benchmarks with varying frame rates.

%% file: sec/3_method.tex
\vspace*{-0.5em}
\section{Method}
\vspace*{-0.5em}
\begin{figure*}[t]
\vspace*{-1em}
  \centering
  \includegraphics[width=0.8\linewidth]{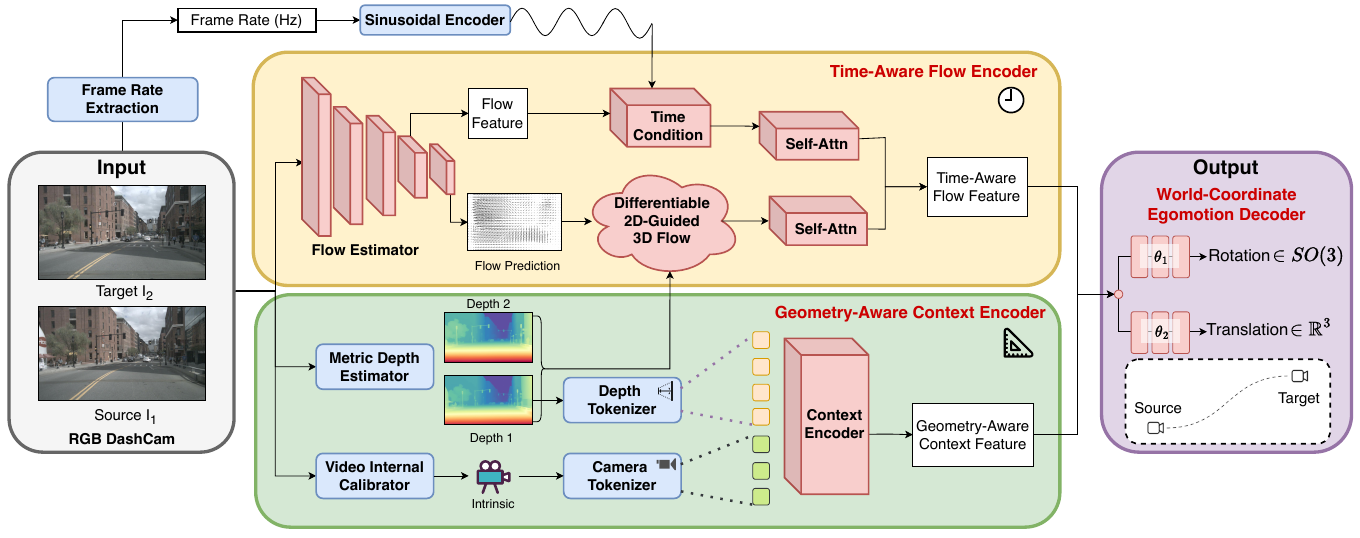}
   \vspace{-0.3cm}
   \caption{\textbf{Overview of \Approach}. We propose a novel temporal-dynamics-informed, geometry-aware visual odometry system. Our method takes consecutive dashcam frames as input and extracts both temporal and geometric representations for robust egomotion estimation. The \textbf{Time-Aware Flow Encoder} (Sec.~\ref{sec:timeawareenc}) leverages a Differentiable 2D-Guided 3D Flow module and time-conditioned embeddings to model motion dynamics across varying observation rates, while the \textbf{Geometry-Aware Context Encoder} (Sec.~\ref{sec:geoawareenc}) incorporates metric depth and intrinsic priors to build a consistent 3D geometry structure of the scene. Finally, the \textbf{World-Coordinate Egomotion Decoder} (Sec.~\ref{sec:worldcoordtraining}) predicts accurate world-coordinate egomotion trajectories from the fused dynamic-geometric representation. 
   }
   \label{fig:architecture}
   \vspace{-0.5cm}
\end{figure*}

In the open world, dashcam videos are recorded across a vast range of devices with diverse lenses, focal lengths, and frame rates, posing significant challenges due to the absence of calibration information. When attempting to reconstruct camera trajectories from such Internet-scale footage, this lack of calibration becomes a fundamental obstacle. Traditional VO methods~\cite{1315094, realtimevo, campos2021orb, wang2021tartanvo, teed2023deep} assume access to known camera intrinsics to recover metric-scale motion, but this assumption breaks down in many downstream tasks, such as processing unconstrained online dashcam videos. \textit{We develop a generalizable VO system capable of extracting temporally metric-consistent geometry from videos captured by uncalibrated cameras with arbitrary frame rates.}

\myheading{Problem Definition:}
Given a dashcam video of $T$ RGB frames $\{\mathbf{I}_i\}_{i=1}^T$, with spatial resolution $(H, W)$, captured under arbitrary frame rates $f$. The task is to recover a consistent world-coordinate ego-motion trajectory without relying on dataset-specific calibration. Following \cite{lai2025zerovo, lai2023xvo}, we focus on the two-frame pose regression setup where we learn a function for mapping two observed image frames $\mathbf{x}_i = \{\mathbf{I}_{i-1}, \mathbf{I}_{i}\}$ to a relative camera pose with real-world scale $\mathbf{y}_i = [\mathbf{R}_i|\mathbf{t}_i] \in \mathrm{SE}(3)$ , where $\mathbf{R}_i\in \mathrm{SO}(3)$ and $\mathbf{t}_i\in \mathbb{R}^3$. 
Our approach removes the dependency on pre-calibrated camera intrinsics. Instead, we infer them directly from visual cues through a lightweight video internal calibration module, which dynamically estimates the intrinsic parameters $\{f_U, f_V, c_U, c_V\}$ (denoted as $\mathbf{K}$) for each video. The estimated intrinsics are jointly leveraged with the optical flow ${\mathbf{F}}$, associated correlation feature ${\mathbf{F}^c}$ under temporal frequency conditions $f$, and the estimated monocular metric depth sequence $\{\mathbf{D}_i\}_{i=1}^T$ to guide the estimation of {\em temporally and spatially} consistent ego-motion. This formulation enables our system to reconstruct the ego-motion of any dashcam video found on the Internet, producing metrically consistent world-coordinate trajectories without ground-truth calibration or sensor-specific assumptions.

\myheading{Overview:} \Approach~, shown in \cref{fig:architecture}, leverages the concept of \textbf{temporal and geometric awareness} to recover accurate vehicle ego-motion. In \cref{sec:timeawareenc}, we inject temporal dynamics into the extracted flow features, aggregating them with differentiable 3D flow to obtain a unified time-aware flow representation. In \cref{sec:geoawareenc}, we introduce a transformer-based encoder that fuses multiple complementary modalities for geometry-aware context, which is finally combined with the time-aware features to regress the relative camera pose. Finally, \cref{sec:worldcoordtraining} describes world-coordinate ego-motion decoder with multi-time-scale training strategy.

\subsection{Time-Aware Flow Encoder}
\label{sec:timeawareenc}
The goal of this encoder is to  extract time-conditioned optical flow features, enabling the model to encode temporal dynamics within the motion representation.
To represent the dynamic motion of the scene, similar to conventional VO frameworks \cite{wang2021tartanvo, lai2023xvo, lai2025zerovo}, we employ a pretrained MaskFlowNet \cite{Zhao_2020_maskflownet} encoder to extract pixel displacement fields from paired image frames. From its intermediate layers, we obtain both the displacement map ${\mathbf{F}}\in\mathbb{R}^{H\times W \times 2}$ representing horizontal and vertical displacement and the correlation feature ${\mathbf{F}}^c \in \mathbb{R}^{H_F\times W_F \times C_F}$, which encodes dense 2D correspondences between the two images. However, the obtained displacement map and its corresponding feature only represent the dense correspondences between frames, without explicitly modeling the underlying motion dynamics. Prior approaches \cite{wang2021tartanvo,lai2023xvo,lai2025zerovo,teed2023deep} are typically trained and evaluated under a fixed temporal sampling rate, an assumption that fails in real-world dashcam data where frame rates vary widely across devices and video compression settings. As a result, these models learn time-agnostic motion features that cannot properly scale or adapt to varying temporal gaps, leading to severe trajectory drift when tested under different frame rates. To address this, we introduce Time Condition Layers that inject frame-rate information into the flow features. This conditioning implicitly encodes pixel-level velocity, thereby enhancing the generalization capability of VO systems in open-world environments.

\myheading{Time Condition Layers:}
To incorporate temporal information into the flow feature, we map the frame rate to a time gap representation $\Delta t = \frac{1}{f}$, where $f$ denotes the frame rate. This time interval $\Delta t \in\mathbb{R}$ is then expanded to high-dimensional embedding using a sinusoidal positional encoding. Specifically, for each frequency scale $\omega_i=\pi\cdot2^i$, for $ i=0,1,2\dots,K-1$, we compute the encoding as:

\vspace{-1em}
\begin{equation}
\begin{split}
\text{PE}(\Delta t) = 
[\,\Delta t,\, 
\sin(\omega_0 \Delta t), \dots, \sin(\omega_{K-1} \Delta t),\\
\cos(\omega_0 \Delta t), \dots, \cos(\omega_{K-1} \Delta t)\,].
\end{split}
\label{eq:pe}
\vspace{-1em}
\end{equation}
This positional encoding injects smooth periodic variations of different frequencies that capture both fine-grained and coarse temporal scales, allowing the network to perceive and generalize across different observation rates. We design two lightweight conditioning layers that take the encoded PE($\Delta t$) as input to modulate the flow feature ${\mathbf{F}}^c$ :
\vspace{-0.5em}
\begin{equation}
\begin{aligned}
\alpha &= \text{Linear}_{\alpha}(\text{PE}(\Delta t)), \\
\beta  &= \text{Linear}_{\beta}(\text{PE}(\Delta t)), \\
\tilde{F^c} &= (1 + \alpha) \odot F^c + \beta,
\end{aligned}
\vspace{-0.5em}
\end{equation}
where $\odot$ denotes the Hadamard product. The adaptive flow features $\tilde{F^c}$ are then processed by four self-attention blocks, which operate on the temporally conditioned feature representation. These attention layers refine the spatial correlations of each pixel, now enriched with temporal cues injected by the Time Condition Layers. 
This enables the network to reason about motion structures and geometric relationships that are consistent with the underlying temporal dynamics, yielding time-aware flow representations that are more robust to varying observation frequencies.

\myheading{Differentiable 2D-Guided 3D Flow:}
We convert the 2D optical flow and per-pixel metric depth into dense 3D motion field using a fully differentiable warping mechanism. First, we back-project each pixel $\mathbf{p}_{1} =  [u,v,1]^\top$ with depth $\mathbf{D}_{1}(u,v)$ into 3D camera coordinates via perspective backprojection $\mathbf{P_{1}}=\mathbf{D}_{1}(u,v)\mathbf{K}^{-1}\mathbf{p}_{1}$. Then, we use the 2D optical flow to warp pixels from the first frame to their sub-pixel locations in the second frame. Given the base pixel grid $(u_1,v_1)$ and the flow field $\mathbf{F} = (\mathbf{F}_x,\mathbf{F}_y)$, we compute the warped coordinate by $(u_1+\mathbf{F}_x, v_1+\mathbf{F}_y)$. We normalize these coordinates and apply bilinear sampling on $\mathbf{D}_{2}$ and filters out points that fall outside the image or have invalid depth, yielding the corresponding depth 
at the flowed locations in a fully differentiable manner. Finally, we back-project the warped pixels to obtain $\mathbf{P_{2}}$ and compute a dense 3D flow $(\mathbf{P_{2}} - \mathbf{P_{1}})$ for all valid pixels. This yields a metric 3D motion field aligned with the camera-1 coordinate system, tightly coupling 2D flow and monocular depth while remaining end-to-end trainable. The resulting 3D flow field encodes per-pixel motion in real-world metric space, providing a geometry-grounded representation of scene dynamics. We further process this 3D flow through four self-attention layers to refine its spatial coherence, and subsequently fuse it with the temporally conditioned flow features produced by the Time Condition Layers. The fused representation, termed the \textbf{Time-Aware Flow Feature}, captures both geometric consistency and motion continuity across temporal dynamics, forming a robust foundation for downstream trajectory estimation.

\subsection{Geometry-Aware Context Encoder}
\label{sec:geoawareenc}
To ensure metric-scale consistency across diverse camera setups, we introduce a Geometry-Aware Context Encoder that explicitly injects depth and camera intrinsic priors into VO pipeline. Traditional monocular VO approaches often rely solely on photometric or flow-based cues \cite{lai2023xvo,wang2017deepvo}, implicitly assuming that appearance information is sufficient to infer motion. However, such representations are inherently scale-ambiguous and fail to generalize when the testing camera differs from those seen during training.
In contrast, our encoder provides a geometry-grounded prior that enables the model to reason about spatial structure and camera projection properties in a unified embedding space.

\myheading{Camera tokenizer:} Different cameras exhibit distinct projection behaviors due to variations in focal length, principal point, and sensor geometry. When training and testing on large-scale dashcam datasets collected from diverse sources, such variations introduce substantial appearance and motion inconsistencies, even for identical underlying 3D motions. To enable camera-aware representation, we leverage a pretrained lightweight internal calibrator WildCamera \cite{zhu2023wildcamera} to infer camera intrinsics from unlabeled dashcam video. We obtain $\mathbf{K}$ simply by averaging the result of \cite{zhu2023wildcamera} across the entire video sequence. Then, we define a normalized intrinsic field:

\vspace{-2em}
\begin{equation}
\begin{aligned}
r(u,v)\propto \mathbf{K}^{-1}[u,v,1]^\top = \left[ \frac{u-c_U}{f_U}, \frac{v-c_V}{f_V}, 1\right ]^\top
\end{aligned}
\label{eq:intrinsiclayer}
\vspace{-0.5em}
\end{equation}
which describes the direction of the 3D viewing ray passing through pixel $(u,v)$ in the camera coordinate system. This intrinsic ray field captures the geometric projection pattern across the image plane: pixels near the optical center correspond to narrower ray cones, while those toward the periphery represent wider projection angles.

\myheading{Depth tokenizer:}  We further incorporate metric depth to provide scene-scale information. Specifically, we leverage a pretrained metric depth estimator Metric3Dv2 \cite{hu2024metric3dv2} to obtain per-pixel depth $\mathbf{D}$. Each ray direction from \cref{eq:intrinsiclayer} is then modulated by the corresponding depth value $\mathbf{M(u,v)} = \mathbf{D}(u,v) \cdot r(u,v)$. This operation projects each pixel onto its corresponding 3D point relative to the camera origin. The resulting spatial distribution $\mathbf{M(u,v)}$ effectively reconstructs the scene structure up to metric scale, capturing both the directionality and magnitude of each visual ray.
Finally, we assemble the token set $\left[ r, \mathbf{M}, \mathbf{D}\right ]$ representing a geometry-aware encoding of the camera projection behavior and scene structure. These tokens are processed by a Geometry-Aware Context Encoder composed of a stack of eight self-attention layers, which learn to correlate the intrinsic and depth priors with the surrounding visual context. This produces a unified geometric embedding that generalizes across uncalibrated dashcam sources, providing a consistent world-coordinate egomotion estimation.

\subsection{World-Coordinate Egomotion Decoder}
\label{sec:worldcoordtraining}
To estimate world-coordinate ego-motion at step $i$, we concatenate Time-Aware Flow Feature $\mathbf{F}^{\text{TA}}$ and Geometry-Aware Context Feature $\mathbf{F}^{\text{GA}}$, which is then mapped to (a) ego translation and (b) rotation by two MLP branches:

\vspace{-0.5em}
\begin{equation}
\begin{aligned}
    \bm{\mathcal{F}} &= f_{\theta_1}([\mathbf{F}^{\mathrm{TA}}, \mathbf{F}^{\mathrm{GA}}]); 
        && \bm{\mathcal{F}} \in \mathbb{R}^{3\times 3} \\
    \mathbf{R}_i &\sim \mathrm{MF}(\bm{\mathcal{F}}); 
        && \mathbf{R}_i \in \mathrm{SO}(3). \\
    \mathbf{t}_i &= f_{\theta_2}([\mathbf{F}^{\mathrm{TA}}, \mathbf{F}^{\mathrm{GA}}]); 
        && \mathbf{t}_i \in \mathbb{R}^3. \\
\end{aligned}
\end{equation}
where $\mathbf{R}_i$ and $\mathbf{t}_i$ are the estimated rotation matrix and translation, respectively, $\bm{\mathcal{F}}$ is the Fisher matrix, and $\mathrm{MF}(\cdot)$ maps Fisher matrix into an element of $\mathrm{SO}(3)$~\cite{mardia2009directional,mohlin2020probabilistic}.
The rotational matrix $\mathbf{R}_i$ has a probabilistic formulation grounded in the Fisher Matrix distribution~\cite{mohlin2020probabilistic,mardia2009directional,lai2023xvo,lai2025zerovo}, which models orientation uncertainty to better model ego-motion variations, as rotations heavily influence the estimated trajectory.
For translation, we employ a metric-scale regression module \cite{wang2021tartanvo} to directly predict world-coordinate displacements, enabling scale-consistent motion recovery.  We optimize the model, $\theta_1,\theta_2$ using MSE loss over $\mathbf{t}_i$, and NLL over $\mathbf{R}_i$. $\mathcal{L} = \lVert \mathbf{t}_i - \hat{\mathbf{t}_i} \rVert_2^2 \;-\; \log \bigl( p(\mathbf{R}_i \mid \boldsymbol{\mathcal{F}} ) \bigr)$.

The temporal dynamics embedded in the Time-Aware Flow Feature $\mathbf{F}^{\mathrm{TA}}$ inform the decoder about the effective temporal scale, improving its ability to model both short- and long-range motions. Since \Approach~is explicitly conditioned on $f=\frac{1}{\Delta t}$, we perform temporal frequency augmentation during training to expose it to multiple time scales. Given an input video recorded at original frame rate $f_{0}$, we generate lower-frequency observations by sub-sampling frames at a factor $k$, producing new training pairs at frame rate $\frac{f_{0}}{k}$ via frame skipping. This allows \Approach~to model motion across short- and long-term temporal scales while enabling the time-aware encoder to adapt its embeddings to different online observation frequencies. To improve training stability, we apply gradient clipping during backpropagation. Using these proposed designs, \Approach~becomes robust to unseen frame-rate variations commonly encountered in Internet-scale dashcam videos.

%% file: sec/4_experiments.tex
\vspace*{-0.5em}
\section{Experiment}
\subsection{Experimental Setup}
\vspace*{-0.5em}
\myheading{Dataset}: Following \cite{lai2023xvo, lai2025zerovo}, we evaluate the generalization capability of \Approach~ under diverse frame rates, camera configurations, and environments using three standard visual-odometry benchmarks: KITTI \cite{geiger2013vision}, nuScenes \cite{caesar2020nuscenes}, and Argoverse 2 \cite{wilson2023argoverse2}.
\textbf{KITTI} contains 11 sequences (00–10) recorded at 10 Hz, each lasting about 4–5 minutes with an average of 4,500 front-view images per scene, providing long-baseline trajectories that stress temporal consistency.
\textbf{nuScenes} comprises 1,000 driving scenes captured at 12 Hz across four regions in Boston and Singapore, each about 20 seconds long with an average of 240 front-view images per scene. It spans diverse conditions, including dense traffic, nighttime, and rainy environments, making it ideal for evaluating temporal adaptability.
\textbf{Argoverse 2} includes 1,000 sequences recorded at 20 Hz, averaging 380 grayscale stereo front-view images per scene, collected from six U.S. cities under varying weather and illumination. This dataset provides a strong test of cross-camera and cross-modality generalization. For a fair comparison with prior works \cite{lai2023xvo,lai2025zerovo}, we downsample all sequences to 160 stereo front-view images (10 Hz). We also report cross-dataset generalization using the 20 Hz results in our ablation study.
Together, these datasets span a broad range of temporal frequencies, camera intrinsics, and scene complexities, forming a rigorous evaluation framework for time-aware open-world visual odometry.

\begin{table*}[!t]
\begin{small}
\centering
\resizebox{16.0cm}{!}{
\begin{tabular}{p{2.9cm}|cccc|cccc|cccc}
\toprule
\multirow{2}{*}{\textbf{Method}} & 
\multicolumn{4}{c|}{\textbf{KITTI 00–10 (10Hz)}} &
\multicolumn{4}{c|}{\textbf{nuScenes (12Hz)}} &
\multicolumn{4}{c}{\textbf{Argoverse 2 (10HZ)}} \\ 
\cline{2-13}
 & $t_{err}$ & $r_{err}$ & ATE & $s_{err}$ 
 & $t_{err}$ & $r_{err}$ & ATE & $s_{err}$
 & $t_{err}$ & $r_{err}$ & ATE & $s_{err}$ \\ 
 \toprule
\rowcolor{gray!30} TartanVO \cite{wang2021tartanvo} & 13.85 & 3.27 & 103.07 & - & 10.27 & 6.35 & 6.26 & - & 11.17 & 5.30 & 7.03 & - \\
\rowcolor{gray!30} DPVO \cite{teed2023deep} & 8.31 & 2.37 & 78.53 & - & 4.34 & 2.85 & 2.66 & - & 2.66 & 1.25 & 1.59 & - \\
\rowcolor{gray!30} ZeroVO$^+$ \cite{lai2025zerovo} & 6.81 & 2.69 & 104.69 & 0.06 & 9.74 & 4.37  & 6.03 & 0.12 & 4.64 & 2.83 & 3.05 & 0.09 \\
\midrule
XVO \cite{lai2023xvo} & 16.82 & 3.84 & 168.43 & 0.17 & 12.75 & 5.11 & 8.30 & 0.16 & 9.13 & 4.86 & 5.70 & 0.12 \\
M+DS \cite{hu2024metric3dv2, teed2021droid} & 14.22 & \textbf{2.72} & 154.77 & 0.09 & 17.08 & \textbf{1.46} & 10.46 & 0.18 & 16.67 & \textbf{1.79} & 8.51 & 0.13 \\
ZeroVO$^{\ddagger}$ \cite{lai2025zerovo} & \underline{8.99} & \underline{2.92} & 123.42 & 0.08 & 12.26 & 5.23 & 8.40 & 0.15 & 8.62 & 4.11 & 5.71 & 0.11\\
\midrule
\textbf{OpenVO$^{\times}$} & 9.02 & 3.32 & \underline{109.01} & \underline{0.06} & \underline{9.57} & \underline{3.71} & \underline{6.13} & \underline{0.11}  & \underline{3.30} & \underline{2.48} & \underline{2.43} & \underline{0.07}\\
\textbf{OpenVO$^{\checkmark}$} & \textbf{8.97} & 3.43 & \textbf{93.23} & \textbf{0.05} & \textbf{9.04} & 3.86 & \textbf{5.91} & \textbf{0.10} & \textbf{3.25} & 2.84 & \textbf{2.39} & \textbf{0.07}\\
\bottomrule
\end{tabular}}
\vspace{-4pt}
\caption{\textbf{Generalizability results.} Comparison with existing methods on 3 large-scale driving benchmarks using standard metrics: translation ($t_{err}$), rotation ($r_{err}$), absolute trajectory (ATE), and scale error ($s_{err}$). \Approach~is \textbf{trained exclusively on nuScenes} and evaluated on KITTI and Argoverse 2 with unseen camera setups, as well as unseen regions of nuScenes, to assess cross-domain generalization.
\colorbox{gray!30}{Shaded rows} denote methods not directly comparable to OpenVO (either rely on ground-truth intrinsics or auxiliary modality). “–” indicates unreported values; $^{+}$ denotes training with additional text guidance and YouTube data, and $^{\ddagger}$ suggests the use of the same Depth \cite{hu2024metric3dv2} and Intrinsic \cite{zhu2023wildcamera} priors from foundation models as ours.
\Approach~has 2 variants: $\checkmark$ with differentiable and $\times$ with non-differentiable 2D-Guided 3D flow. 
The \textbf{best} and \underline{second best} results are marked as \textbf{bold} and \underline{underline} among intrinsic-free, vision-only models.
}
\end{small}
\label{tab:zero-shot}
\vspace{-10pt}
\end{table*}

\begin{figure*}[t]
  \centering
  \includegraphics[width=0.9\linewidth]{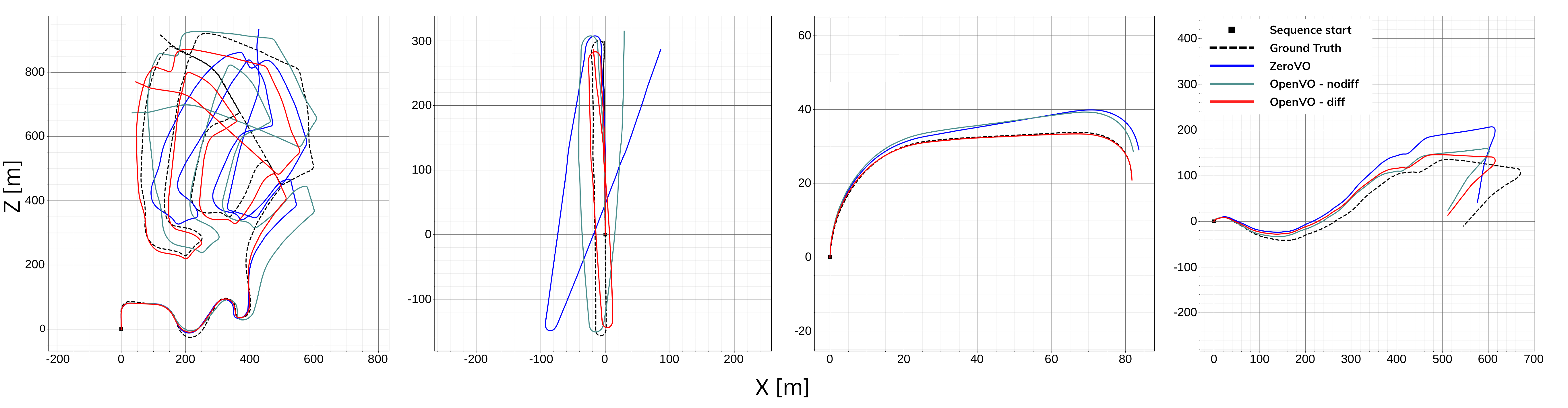}
   \vspace{-10pt}
   \caption{\textbf{Qualitative results}. We present trajectory prediction results on the KITTI and nuScenes datasets. Compared to ZeroVO$^{\ddagger}$, both variants of our method --- differentiable (\textbf{OpenVO-diff}) and non-differentiable variants (\textbf{OpenVO-nodiff}) of our 2D-guided 3D flow --- achieve higher trajectory prediction accuracy and consistency, surpassing the current state-of-the-art.}
   \label{fig:quali}
   \vspace{-10pt}
\end{figure*}

\begin{table}[!t]
\centering
\setlength\tabcolsep{3pt} 
\renewcommand{\arraystretch}{1.0}

\begin{tabular}{l|cccc|cccc}
\toprule
\multirow{2}{*}{\textbf{\small Setting}} & 
\multicolumn{4}{c|}{\textbf{\small KITTI 00–10 (10Hz)}} &
\multicolumn{4}{c}{\textbf{\small nuScenes (12Hz)}} \\ 
\cline{2-9}
 & $t_{err}$ & $r_{err}$ & ATE & $s_{err}$ 
 & $t_{err}$ & $r_{err}$ & ATE & $s_{err}$ \\ 
\midrule
\textbf{$K=2$} & 9.68 & 3.55 & 108.48 & 0.07 & 9.47 & 3.90 & 6.63 & 0.11 \\
\textbf{$K=4$} & 9.43 & 3.54 & 101.53 & 0.06 & 9.95 & \textbf{3.60} & 6.74 & 0.12 \\
\textbf{$K=8$} & \textbf{8.97} & 3.43 & \textbf{93.23} & \textbf{0.05} & \textbf{9.04} & 3.86 & \textbf{5.91} & \textbf{0.10} \\
\textbf{$K=16$} & 9.18 & 3.72 & 95.11 & 0.06 & 9.85 & 6.60 & 6.30 & \textbf{0.10} \\
\textbf{$K=32$} & \textbf{8.97} & \textbf{3.07} & 114.22 & 0.08 & 10.21 & 5.11 & 7.03 & 0.12 \\
\textbf{$K=64$} & 10.24 & 4.00 & 119.23 & 0.07 & 11.15 & 4.44 & 6.92 & 0.13 \\
\bottomrule
\end{tabular}
\vspace{-4pt}
\caption{\textbf{Ablation on Time Encoding's Token Size ($K$).}}
\label{tab:ablate_TFS}
\vspace{-15pt}
\end{table}

\begin{table}[!t]
\centering
\setlength\tabcolsep{2pt} 
\renewcommand{\arraystretch}{1.0}
\begin{tabular}{l|cccc|cccc}
\toprule
\multirow{2}{*}{\textbf{\small Setting}} & 
\multicolumn{4}{c|}{\textbf{\small KITTI 00–10 (10Hz)}} &
\multicolumn{4}{c}{\textbf{\small nuScenes (12Hz)}} \\ 
\cline{2-9}
 & $t_{err}$ & $r_{err}$ & ATE & $s_{err}$ 
 & $t_{err}$ & $r_{err}$ & ATE & $s_{err}$ \\ 
\midrule
\rowcolor{gray!30}\{12/6/4\} & 10.24 & 4.24 & 152.42 & 0.09  & 9.38 & 4.02 & 6.15 & 0.11 \\
\midrule
\{12\}$^{*}$ & 12.89 & 4.40 & 213.76 & 0.09  & 10.23  & 4.91 & 6.35 & 0.12 \\
\{12/6\} & \textbf{8.64} & \textbf{2.91} & 117.58 & 0.07 & \textbf{8.08} & \textbf{3.40} & \textbf{5.27} & \textbf{0.10} \\
\{12/6/4\} & 8.97 & 3.43 & \textbf{93.23} & \textbf{0.05} & 9.04 & 3.86 & 5.91 & \textbf{0.10} \\
\{12/6/4/3\} & 9.43 & 3.57 & 97.21 & 0.06 & 9.06 & 4.21 & 6.02 & 0.11 \\
\bottomrule
\end{tabular}
\vspace{-4pt}
\caption{\textbf{Ablation on Training Observation Frequency.} \colorbox{gray!30}{Shaded row}: \Approach~trained \textbf{\textit{without}} proposed Time Condition Layers; \colorbox{white!30}{Other rows}: \Approach~trained \textbf{\textit{with}} proposed Time Condition Layers.  $^{*}$ indicates that \Approach~uses 12 Hz as the fixed inference temporal condition, regardless of the dataset.}
\label{tab:ablate_FT}
\vspace{-16pt}
\end{table}

\begin{table}[!t]
\centering
\setlength\tabcolsep{1.5pt} 
\renewcommand{\arraystretch}{1.0}
\begin{tabular}{l|ccc|ccc}
\toprule
\multirow{2}{*}{\textbf{Setting}} & 
\multicolumn{3}{c|}{\textbf{OpenVO}} &
\multicolumn{3}{c}{\textbf{ZeroVO}$^{\ddagger}$ \cite{lai2025zerovo}} \\ 
\cline{2-7}
 & $t_{err}$ & $r_{err}$ & ATE  
 & $t_{err}$ & $r_{err}$ & ATE  \\ 
\midrule
\textit{KITTI (2.5Hz)} & \cellcolor{green!10}{28.83} & \cellcolor{green!10}{8.22} & \cellcolor{green!10}{368.47} & 52.74 & 17.46 & 553.52   \\
\textit{KITTI (5Hz)} & \cellcolor{green!10}{12.06} & \cellcolor{green!10}{3.77} & \cellcolor{green!10}{132.72}  & 28.62 & 8.78 & 387.10  \\
\midrule
\textit{nuScenes (3Hz)} & \cellcolor{green!10}{12.73} & \cellcolor{green!10}{3.71} & \cellcolor{green!10}{7.60} &  25.75 & 8.30 & 14.64   \\
\textit{nuScenes (4Hz)} & \cellcolor{green!10}{8.14} & \cellcolor{green!10}{3.24} & \cellcolor{green!10}{5.00} & 15.49  & 6.08 & 9.33    \\
\textit{nuScenes (6Hz)} & \cellcolor{green!10}{9.18} & \cellcolor{green!10}{4.05} & \cellcolor{green!10}{6.07} &  31.24 & 10.30 & 21.55   \\
\midrule
\textit{Argoverse 2 (4Hz)} & \cellcolor{green!10}{3.63} & \cellcolor{green!10}{1.72} & \cellcolor{green!10}{2.09} & 25.93 & 19.62 & 28.38   \\
\textit{Argoverse 2 (5Hz)} & \cellcolor{green!10}{4.40} & \cellcolor{green!10}{1.71} & \cellcolor{green!10}{2.47} & 31.79 & 16.20 & 26.83   \\
\textit{Argoverse 2 (20Hz)} &\cellcolor{green!10}{9.79}  & \cellcolor{green!10}{4.72} & \cellcolor{green!10}{6.47} & 61.61 & 27.38 & 36.14    \\

\bottomrule
\end{tabular}
\vspace{-8pt}
\caption{\textbf{
    Ablation on Varying Inference Observation Rates.} $^{\ddagger}$ indicates ZeroVO uses the same Metric Depth \cite{hu2024metric3dv2} and Intrinsic \cite{zhu2023wildcamera} priors from foundation models as \Approach. 
    \colorbox{green!5}{Green cells} denote ~\Approach~{\em (left)} improved performance of over existing method~{\em (right)}.
    Our~\Approach~is {\em significantly more robust to observation rate variations, consistently reducing VO errors} by {\bf 46\%} to {\bf 92\%} across different metrics and benchmark settings.
}
\label{tab:ablate_diversetimegap}
\vspace{-20pt}
\end{table}

\myheading{Evaluation metrics:}
We evaluate our method using three standard metrics: Translation Error ($t_{err}$), Rotation Error ($r_{err}$), and Absolute Trajectory Error (ATE), along with a recently introduced Scale Error ($s_{err}$) \cite{lai2023xvo,lai2025zerovo}.
The $t_{err}$ measures the average translational drift (in \%) between the predicted and ground-truth trajectories, reflecting how accurately the model estimates displacement over time. The $r_{err}$ quantifies the angular deviation (in $^\circ$/100\,m) between predicted and true orientations, indicating how well rotational motion is captured. The ATE (in meters) assesses global trajectory consistency by aligning the predicted and ground-truth paths and computing the root-mean-square (RMS) positional difference, revealing how local errors accumulate over time. Finally, the $s_{err}$ measures the relative discrepancy in translation magnitudes, evaluating how faithfully the model preserves metric scale. It is computed as the mean difference between L2 norms of translation vectors across timesteps. Our metric \& eval protocol follow ZeroVO, using \textbf{only first-frame alignment}, i.e., we set the first predicted pose to match the ground-truth pose and apply \textbf{no SE(3) or Sim(3)} alignment afterward. The models must predict metric-scale poses directly and no GT alignment applied for a fair comparison.

\myheading{Setup details:}
Following \cite{lai2025zerovo}, we train \Approach~on the Singapore-OneNorth subset (183 scenes) and evaluate it on unseen regions of nuScenes, as well as KITTI and Argoverse 2 for cross-domain generalization. Testing is performed without access to camera intrinsics. For WildCamera, we use the released nine-dataset weights to estimate intrinsics, and for Metric3Dv2, we adopt DINO2reg-ViT-Large with WildCamera-predicted intrinsics for metric depth estimation. \Approach~ is trained for 285.6k iterations (batch = 16) with Random Crop \& Resize \cite{wang2021tartanvo} and Horizontal Flip (seed = 2023), using $K=8$ and temporal-frequency augmentation at 4, 6, and 12 Hz. Each experiment takes about 96 GPU hours on an A6000. Unlike \cite{lai2025zerovo,lai2023xvo}, we use no additional YouTube data to ensure a fair comparison.
\vspace*{-0.5em}
\subsection{Comparison to prior work}
\vspace*{-0.5em}
\myheading{Quantitative comparison:}
The quantitative evaluation of \Approach~is summarized in \cref{tab:zero-shot}. We compare \Approach~against prior state-of-the-art methods, including XVO \cite{lai2023xvo}, ZeroVO \cite{lai2025zerovo}, and the SLAM-based baseline Metric3Dv2~\cite{hu2024metric3dv2} + DroidSLAM~\cite{teed2021droid} (M+DS) introduced in \cite{lai2025zerovo}. We conduct zero-shot testing on \textbf{KITTI 00–10 (10 Hz)} under unknown camera parameters. \Approach~achieves a global trajectory ATE of 93.23, outperforming ZeroVO$^{\ddagger}$ by 24\% while using the same Depth + Intrinsic priors, and even surpassing ZeroVO$^{+}$ trained with additional text guidance and preprocessed YouTube data.
In terms of real-world scale recovery, our method attains state-of-the-art performance, demonstrating strong generalization to unseen real-world videos. However, slightly higher local trajectory segment errors in $t_{err}$ and $r_{err}$ arise from training with mixed temporal frequencies, which can lead to inconsistent parameter updates despite having applied gradient clipping. This limitation can be mitigated through adaptive temporal sampling, which we leave for future investigation. For \textbf{nuScenes (12 Hz)}, we evaluate on unseen regions and achieve state-of-the-art results across all four metrics. \Approach~achieves an ATE of 5.91, effectively capturing scene geometry from limited input and outperforming methods with additional supervision. Under the same setting, both $t_{err}$ and $r_{err}$ also surpass the current state-of-the-art, confirming superior convergence. For \textbf{Argoverse 2 (10 Hz)}, we achieve superior performance across all four metrics, even surpassing the YouTube text-guidance self-supervised model from ZeroVO$^{+}$. Since the trajectories in Argoverse 2 are generally less complex than those in KITTI, most methods -- including ours, attain an ATE below 10.
Across all three datasets, our differentiable 2D-Guided 3D Flow variant yields more robust trajectory predictions. The differentiable design enables absorption of dynamic 4D geometric cues from foundation priors while keeping the pipeline end-to-end trainable. Even without differentiability, \Approach$^{\times}$~achieves state-of-the-art results, highlighting its generalization to unseen dashcam videos through strong foundation priors and context encoding.

\vspace*{-0.25em}
\myheading{Qualitative comparison:} We visualize the X-Z projection qualitative trajectory predictions of ZeroVO$^{\ddagger}$ and the two variants of our \Approach~in \cref{fig:quali}. The differentiable version produces trajectories that align more closely with the ground truth, especially in complex or long-range scenes, highlighting the benefit of gradient-consistent 3D flow modeling and improved geometric stability. \textit{Furthermore, we showcase the versatility of \Approach~across diverse autonomous driving tasks and applications in the Supp. Mat.}

\vspace*{-0.25em}
\subsection{Ablation Studies}
\vspace*{-0.25em}
To validate design choices of our method, we conduct a series of ablation studies on different configuration setups.

\vspace*{-0.5em}
\myheading{Study on temporal frequency size} is in \cref{tab:ablate_TFS}. The size of the positional encoding from \cref{eq:pe} is $(1+2K)$. We observe that {\em setting $K=8$ achieves the best performance by providing a balanced range of temporal frequencies} that capture both fine and coarse-scale motion dynamics.
Using lower $K$ limits temporal expressiveness, causing the model to under-fit variations in frame rates, while excessively large $K$ introduces unstable high-frequency oscillations. Therefore, $K=8$ offers an effective trade-off between representational richness and training stability. 

\vspace*{-0.5em}
\myheading{Effect of multi-time-scale training} demonstrates in \cref{tab:ablate_FT}. 
Training solely at 12 Hz (second row) restricts the model to a fixed temporal dynamic, leading to degraded performance on both datasets. On KITTI, testing under the same 12 Hz condition since the model never encountered 10 Hz encodings, resulting in a large ATE of 213.76. Nonetheless, the 12 Hz-only model surpasses ZeroVO$^{\ddagger}$ on nuScenes, validating the effectiveness of our differentiable design and geometry-aware context encoder. Incorporating 6 Hz samples (third row) enables the model to exploit time-conditioned features, yielding consistent gains across datasets. Further adding 4 Hz (fourth row) improves ATEs to 93.23 and 5.91 on KITTI and nuScenes, though local consistency slightly decreases due to over-diverse temporal patterns -- suggesting adaptive sampling as a remedy. Adding another frequency (last row) yields asymptotic improvement, indicating convergence in temporal generalization.
One might argue that mixing multi-frequency data alone could allow the model to implicitly learn temporal relationships. However, {\em when we remove the Time Condition Layers (first row), forcing the network to self-discover temporal dynamics, performance drops notably} on KITTI as it becomes confused by inconsistent motion patterns -- {\em underscoring the necessity of our temporal awareness for robust multi-time-scale learning}.

\vspace*{-0.5em}
\myheading{Analysis of varying temporal gap} is presented in \cref{tab:ablate_diversetimegap}. To set up this configuration, for each dashcam video, we subsample the original framerate by a factor of $k$ that is, we skip $(k-1)$ frames between consecutive inputs and apply the same subsampling to the corresponding ground-truth camera poses. The original capture rates are 10 Hz for KITTI, 12 Hz for nuScenes, and 20 Hz for Argoverse 2.
We compare our OpenVO with the conventional ZeroVO$^{\ddagger}$. As shown, ZeroVO$^{\ddagger}$ exhibits substantial degradation when evaluated under varying frame rates, highlighting its inability to adapt to changes in temporal dynamics. This demonstrates that {\em modeling temporal dynamics is critical for robust egomotion estimation under real-world video conditions}, where frame rates are often inconsistent.

%% file: sec/5_conclusion.tex
\section{Conclusion}

We present \Approach, a real-world–scale egomotion estimation framework that models temporal dynamics and geometric consistency from uncalibrated dashcam videos. 

\vspace{-0.5em}
\myheading{Limitation:} Despite its generalizability, Depth Estimator and Internal Calibrator in ~\Approach~function separately, so errors in metric depth or intrinsics may propagate to final VO results. However, this is not unique to~\Approach~but is a common limitation across intrinsic-free VO literature. Moreover, the current multi-time-scale settings are set empirically. A more principled strategy, where the model adaptively adjusts the sampling frequency on the fly in a synergistic manner, would be an interesting future direction.

\noindent
{\bf Acknowledgments:} This project is supported in part by
Barry Mersky and Capital One Endowed Professorships.

%% file: supp_mat.tex






\maketitlesupplementary

In this supplementary material, we first provide a more detailed implementation of our Differentiable 2D-Guided 3D Flow in \cref{sec:diffflow}. We then present additional training and evaluation analysis in \cref{sec:analysis}. Furthermore, we outline potential applications enabled by OpenVO in \cref{sec:application}. Finally, we include additional qualitative results in \cref{sec:qualiresults}, demonstrating the robustness and superior trajectory reconstruction capabilities of OpenVO across diverse scenarios.

\section{Differentiable 2D-Guided 3D Flow}
\label{sec:diffflow}

Given two consecutive frames with metric depth and a dense optical flow field, our differentiable 2D-guided 3D flow layer computes a per-pixel 3D flow in the coordinate system of the first camera. Let $\mathbf{D}_1$, $\mathbf{D}_2 \in \mathbb{R}^{H\times W}$ denote the metric depth maps at times $t_0$ and $t_1$, respectively, expressed in meters. The camera intrinsics are parameterized as:

\begin{equation}
K = \begin{bmatrix}
f_x & 0   & c_x \\
0   & f_y & c_y \\
0   & 0   & 1
\end{bmatrix}
\end{equation}
and the optical flow from $t_0\rightarrow t_1$ is given by:

\begin{equation}
\mathbf{F}(u,v) = \begin{bmatrix}
\Delta u(u,v) \\
\Delta v(u,v)\\
\end{bmatrix},
\qquad
\mathbf{F} \in \mathbb{R}^{H \times W \times 2}
\end{equation}
where $(u,v)$ indexes pixel coordinates. We back-project every pixel $(u,v)$ by constructing the homogeneous image coordinate $\mathbf{p}$ into 3D using the depth $\mathbf{D}_1(u,v)$

\begin{equation}
\mathbf{P}_1(u,v)
= \mathbf{D}_1(u,v)\, K^{-1} \mathbf{p}
= \begin{bmatrix}
X_1(u,v) \\
Y_1(u,v) \\
Z_1(u,v)
\end{bmatrix}
\in \mathbb{R}^3
\label{eq:back-proj}
\end{equation}
In coordinates, this is equivalent to:

\begin{equation}
\begin{bmatrix}
X_1(u,v) \\
Y_1(u,v) \\
Z_1(u,v)
\end{bmatrix}
=
\begin{bmatrix}
\displaystyle \frac{(u - c_x)\, \mathbf{D}_1(u,v)}{f_x} \\
\displaystyle \frac{(v - c_y)\, \mathbf{D}_1(u,v)}{f_y} \\
\displaystyle \mathbf{D}_1(u,v)
\end{bmatrix}
\end{equation}

To establish dense correspondences between the two consecutive frames, we warp each pixel $(u,v)$ from time $t_0$ to its estimated sub-pixel location at time $t_1$ using the optical flow field. The warped pixel coordinates are computed as:

\begin{equation}
u'=u+\Delta u(u,v), \qquad v'=v+\Delta v(u,v)
\end{equation}
This operation allows the model to reason about motion at a finer level than integer pixel shifts, which is crucial for handling small object motions, rolling-shutter distortions, and variations in frame rates.
Since the predicted coordinates $(u', v')$ are generally non-integer and may lie between pixel centers, we perform differentiable sampling of the target depth map $\mathbf{D}_2$. 
We define the final sampling grid:
\begin{equation}
\mathbf{g}(u,v) = \begin{bmatrix}
{u}'(u,v) \\
{v}'(u,v)
\end{bmatrix}
\end{equation}
We declare an operation $\sigma(\cdot)$ which takes metric depth $\mathbf{D}_2$ and sampling grid $\mathbf{g}$ as input and outputs sampled metric depth $\mathbf{\tilde{D}}_2(u,v)$ as:
\begin{equation}
\mathbf{D}_2(u',v')
= \sum_{i \in \{0,1\}} \sum_{j \in \{0,1\}}
w_{ij}\, \mathbf{D}_2(u_i, v_j)
\end{equation}
where pixel neighbors are:
\begin{equation}
u_i = \lfloor u' \rfloor + i,\quad
v_j = \lfloor v' \rfloor + j,\quad
i,j \in \{0,1\}
\end{equation}
and bilinear interpolation weights are:
\begin{equation}
\gamma = u' - u_0,\qquad
\psi = v' - v_0,
\end{equation}

\begin{equation}
w =
\begin{bmatrix}
(1-\gamma)(1-\psi) & (1-\gamma)\psi \\
\gamma(1-\psi) & \gamma\psi
\end{bmatrix}
\end{equation}
We obtain $\mathbf{\tilde{D}_2} = \sigma(\mathbf{D}_2, \mathbf{g})$, then following ~\cref{eq:back-proj}, we back-project $\mathbf{\tilde{D}}_2$ into 3D at time $t_1$ to obtain $\mathbf{P}_2(u,v)$. Finally, we compute a dense 3D flow for every pixel:
\begin{equation}
    \mathbf{S}(u,v) = (\mathbf{P}_2 - \mathbf{P}_1)
\end{equation}
The entire pipeline used to obtain the warped depth
$\mathbf{\tilde{D}}_2(u,v)$ is fully differentiable with respect to the depth map, the
warped coordinates $(u',v')$, and the predicted optical flow
$(\Delta u, \Delta v)$. Since the warped position satisfies
$u' = u + \Delta u$ and $v' = v + \Delta v$, gradients propagate through the
sampling process via the chain rule. For example:
\begin{equation}
\frac{\partial \mathbf{\tilde{D}}_2}{\partial \Delta u}
= \frac{\partial \mathbf{\tilde{D}}_2}{\partial u'} \cdot
  \frac{\partial u'}{\partial \Delta u}
= \frac{\partial \mathbf{\tilde{D}}_2}{\partial u'}
\end{equation}
This smooth interpolation ensures stable gradient flow across sub-pixel
displacements, enabling reliable end-to-end optimization for Time-Aware Flow Encoder block.

To ensure that only geometrically valid correspondences contribute to the loss, we construct a binary validity mask
\begin{equation}
\begin{aligned}
\mathbf{m}(u,v) = \mathbf{1}\big[\,
&0 \le u' \le W-1,\;\;
  0 \le v' \le H-1,\\
&\mathbf{D}_1(u,v) > 0,\;\;
 \mathbf{\tilde{D}}_2(u,v) > 0
\big].
\end{aligned}
\end{equation}
where $\mathbf{1}$ is an indicator function and apply it element-wise to the 3D flow:
\begin{equation}
    \mathbf{\tilde{S}}(u,v) = \mathbf{m}(u,v)\mathbf{S}(u,v) 
\end{equation}
The mask 
$\mathbf{m}$ is a non-learned, piecewise-constant tensor computed from geometric constraints; during backpropagation we do not differentiate through the indicator function, so the operation remains differentiable w.r.t. 
$\mathbf{S}$, and gradients are simply zeroed out at invalid pixels.

\section{Analysis}
\label{sec:analysis}

\begin{table}[t]
\small
\setlength{\tabcolsep}{5pt}
\centering
\begin{tabular}{lcccc}
\toprule
\textbf{Method} & \textbf{\# params (M)} & \textbf{Runtime (s)} \\
\midrule
WildCamera \cite{zhu2023wildcamera}& 270.447 & 41.906\\
Metric3Dv2 \cite{hu2024metric3dv2}& 411.941 & 180.752\\
\bottomrule
\end{tabular}
\vspace{-4pt}
\caption{\textbf{Expected latency of foundation priors.}}
\label{tab:runtime_estimator}
\vspace{-10pt}
\end{table}

\begin{table}[t]
\small
\setlength{\tabcolsep}{5pt}
\centering
\begin{tabular}{lcccc}
\toprule
\textbf{Method} & \textbf{\# params (M)} & \textbf{Runtime (s)} \\
\midrule
\rowcolor{red!20}Flow Estimator& 20.655&3.089 \\
\rowcolor{red!20}Time Condition Layers & 0.003673 & 0.0101\\
\rowcolor{red!20}Diff. 2D-Guided 3D Flow & - & 0.0204 \\
\rowcolor{red!20}Self-Attn & 174.695& 0.0339 \\
\midrule
\rowcolor{green!20}Context Encoder&56.702 & 0.0129\\
\midrule
\rowcolor{purple!20}Egomotion Decoder&  29.988&0.4368 \\
\bottomrule
\end{tabular}
\vspace{-4pt}
\caption{\textbf{Expected latency of our OpenVO.}~\colorbox{red!20}{Red rows} indicate Time-Aware Flow Encoder block. \colorbox{green!20}{Green row} indicates Geometry-Aware Context Encoder block. \colorbox{purple!20}{Purple row} indicates Egomotion decoder block.}
\label{tab:runtime_openvo}
\vspace{-10pt}
\end{table}

\begin{table}[!t]
\centering
\setlength\tabcolsep{2pt}
\renewcommand{\arraystretch}{1.0}
\begin{tabular}{l|cccc|cccc}
\toprule
\multirow{2}{*}{\textbf{\small Setting}} & 
\multicolumn{4}{c|}{\textbf{\small KITTI 00–10 (10Hz)}} &
\multicolumn{4}{c}{\textbf{\small nuScenes (12Hz)}} \\ 
\cline{2-9}
 & $t_{err}$ & $r_{err}$ & ATE & $s_{err}$ 
 & $t_{err}$ & $r_{err}$ & ATE & $s_{err}$ \\ 
\midrule
Ours--Oracle
&  7.22 & 2.75 & 90.85 & 0.05
& 6.41 & 3.44 &4.24 &0.09 \\
\bottomrule
\end{tabular}
\vspace{-4pt}
\caption{\textbf{Oracle testing.} 
We equip OpenVO with ground-truth intrinsic parameters to isolate errors from calibration, revealing the upper-bound performance under perfect camera parameters estimation.}
\label{tab:oracle}
\vspace{-10pt}
\end{table}

\begin{table}[!t]
\setlength{\tabcolsep}{2pt}
\centering
\begin{tabular}{l|c|ccc}
\toprule
\textbf{Time-Conditioning Layers} & \textbf{P.E.}   & $t_{err}$ & $r_{err}$ & ATE  \\
\midrule
 Cross-Attn (PE($\Delta t$) in Eq.1) & \Checkmark  & 12.74 & 4.22 & 155.30   \\
 Cross-Attn (single scalar $\Delta t$) & \XSolidBrush  & 14.74 & 4.78 & 180.16  \\
\bottomrule
\end{tabular}
\caption{\textbf{Time-Conditioning Layers design.} Using Cross-Attention layers as alternatives to the design of Time-Conditioning Layers and the use of our proposed positional encoding (P.E.).}
\label{tab:cross-pe}
\vspace{-10pt}
\end{table}

\begin{table}[!t]
\setlength{\tabcolsep}{3pt}
\centering
\begin{tabular}{l|ccc|ccc}
\toprule
\textbf{Ablation} & \textbf{P.E.} & \textbf{Flow} & \textbf{Geometry}  & $t_{err}$ & $r_{err}$ & ATE  \\
\midrule
OpenVO & \Checkmark & \Checkmark & \XSolidBrush & 17.12&4.42&158.67   \\
OpenVO & \Checkmark & \XSolidBrush & \Checkmark & - & - & -   \\
OpenVO & \XSolidBrush & \Checkmark & \Checkmark & 9.59 & 3.54 & 117.50   \\
\bottomrule
\end{tabular}
\caption{\textbf{Time-Conditioning Layers design.} Using Cross-Attention layers as alternatives to the design of Time-Conditioning Layers and the use of our proposed positional encoding (P.E.).}
\label{tab:feat_ablate}
\vspace{-10pt}
\end{table}

\begin{table}[!t]
\setlength{\tabcolsep}{3pt}
\centering
\begin{tabular}{l|ccc}
\toprule
\textbf{Methods}   & $t_{err}$ & $r_{err}$ & ATE  \\
\midrule
\textbf{PnP}+Metric3Dv2+MaskFlow&  54.38 & 2.81 & 302.74 \\
\textbf{DroidCalib} \cite{hagemann2023deep}: no Sim(3) (\underline{fair}) & 69.16 & 0.625 & 382.65\\
\textbf{DroidCalib} \cite{hagemann2023deep}: w/ Sim(3) (\underline{unfair})  & 11.01 & 0.625 & 73.18\\
\textbf{ViPE} \cite{huang2025vipe}: no Sim(3) (\underline{fair}) & 13.29 & 0.721 & 137.15\\
\textbf{ViPE} \cite{huang2025vipe}: w/ Sim(3) (\underline{unfair}) & 10.15 & 0.721 & 72.73\\
\bottomrule
\end{tabular}
\caption{\textbf{Geometry-based and optimization-based methods.}}
\label{tab:geo-method}
\vspace{-10pt}
\end{table}

\myheading{Motivation:} $\Delta t$ is a conditioning signal indicating the temporal stride of the input pair, serving as a proxy for providing temporal context so the network can produce stable pose increments across different sampling intervals as shown in Tab. 4 (main paper). The purpose of conditioning is not to deterministically scale the predicted pose or to ``force" a particular motion magnitude, but to {{\em adapt to a wide range of motion magnitudes}} -- a behavior that prior methods cannot achieve. Without TCL, the model must self-discover motion patterns without any explicit temporal modeling, leading to suboptimal learning as the single flow encoder backbone is overloaded with multiple conflicting objectives (differentiating small, medium, and large motion magnitudes). Our TCL with the proposed P.E. provides the necessary temporal bridge, stabilizing and improving learning. Unlike prior work, our Context Encoder jointly encodes image–depth inputs and camera parameters (L.361), {enabling the model to be \textit{camera-aware} and generalize effectively to \textbf{any unseen} camera settings during testing}. 

\myheading{Latency:} We report the number of parameters and the total running time for each component of our OpenVO.
For the internal calibrator WildCamera \cite{zhu2023wildcamera} and metric depth estimator Metric3Dv2 \cite{hu2024metric3dv2}, we report the expected runtime for a \textit{single process on 320 image frames} of Argoverse 2 \cite{wilson2021argoverse} in ~\cref{tab:runtime_estimator}. For OpenVO, we present the number of parameters and runtime for each components for a single forward for batch of 16 pairs of images in \cref{tab:runtime_openvo}. Because the flow encoder performs convolution operations over full-resolution image grids, while the self-attention blocks operate on patch-level representations, the two modules exhibit distinct runtime behaviors.

\myheading{Oracle Test:} Given that accurate camera intrinsics are pivotal to our pipeline, we further examine an oracle scenario where the method is supplied with ground-truth intrinsic parameters. We conduct these oracle evaluations on nuScenes~\cite{caesar2020nuscenes} and KITTI~\cite{geiger2013kitti} to quantify the potential performance gains obtainable under perfect calibration in~\cref{tab:oracle}.

\myheading{Ablation on Time-Conditioning Layers:} We further replace Cross-Attention as the Time-Conditioning Layer (TCL) and ablate our positional encoding in \cref{tab:cross-pe}.
When disregard our proposed P.E., we use a single scalar $\Delta t$ as a temporal condition. The results validate the effectiveness of our design, potentially motivating the 3D/4D reconstruction communities to leverage temporal dynamics for computer vision problems.

\myheading{Ablation on the proposed components:} We provide an ablation on the feature stream of OpenVO in \cref{tab:feat_ablate}. Without the flow branch, we cannot estimate the geometry difference between two consecutive frames. OpenVO without explicit geometry (depth) modeling and camera information suffers from generalization with $t_{err}$ of 17.12.

\myheading{Comparison with geometric-based and optimization-based methods:} Traditional closed-world monocular VO/SLAM approaches with known intrinsics, backend optimization \& loop closure, are evaluated with known GT poses alignment. Our OpenVO is evaluated under a strictly harder setting: unknown intrinsics, no GT alignment, no loop closure, and \textit{\textbf{zero-shot cross-dataset generalization}}. We showcase some state-of-the-art (2026-March) and traditional geometry-based methods on visual odometry in \cref{tab:geo-method}. Our OpenVO still achieves state-of-the-art results on reconstructing real-world scale egomotion without the need of alignment.

\section{Global High-Definition (HD) Semantic Maps Reconstruction}
\label{sec:application}

HD maps play a critical role in autonomous driving and 3D scene understanding, providing precise lane geometry, road topology, traffic elements, and other structural cues that are necessary for enhanced situational awareness and safer control and navigation~\cite{liu2023vectormapnet, shi2024globalmapnet,yuan2024streammapnet}. Such maps enable downstream tasks—including planning, motion prediction, navigation, and safety validation—to operate with spatial awareness and robust scene priors. However, HD maps are expensive to produce, typically requiring a combination of LiDAR sensors, human annotation, and specialized mapping vehicles. This motivates reconstructing HD maps directly from onboard sensors like cameras as a more scalable alternative. Yet this approach faces substantial technical challenges, including heavy dependency on ego-motion, occlusions, sensor noise, and complex 3D scene structure, which together can lead to compounding errors in the reconstructed maps over the horizon, especially under uncalibrated monocular observation under varying frame rates. Integrating OpenVO into the mapping pipeline helps mitigate these issues by providing accurate frame-to-frame poses that register local map predictions into a coherent global representation in world coordinates. This combination compensates for camera motion and enables consistent multi-view fusion, allowing scalable, camera-only reconstruction of high-quality HD semantic maps.

\begin{figure*}[t]
\vspace*{-1em}
  \centering
  \includegraphics[width=0.95\linewidth]{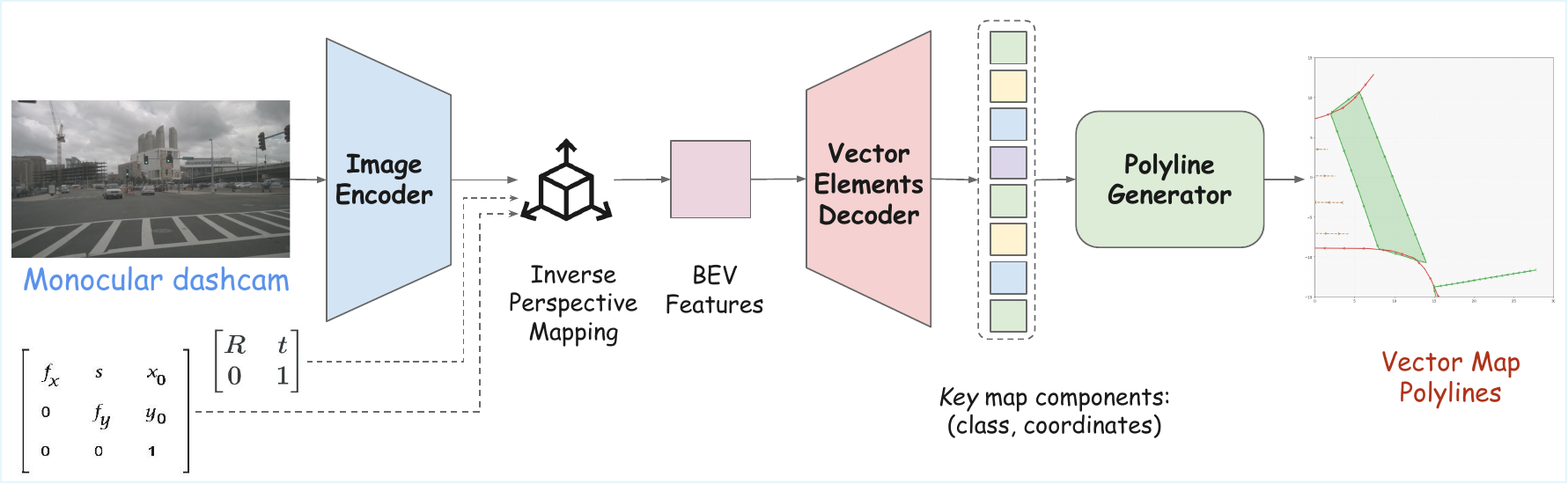}
   \vspace{-0.3cm}
   \caption{\textbf{Modified VectorMapNet \cite{liu2023vectormapnet}.} A front-view input image is first processed by an image encoder to extract semantic and geometric features. These features are then lifted into a bird’s-eye-view (BEV) representation using inverse perspective mapping, which leverages the camera’s intrinsic and extrinsic parameters from \textbf{OpenVO} to geometrically project image features onto the ground plane. The resulting BEV feature map is fed into the Vector Map Decoder, which predicts structured map elements in an intermediate representation consisting of key components such as polyline classes, control points, and geometric attributes. Finally, a polyline generator converts these decoded components into continuous vectorized map elements, such as lane boundaries, road dividers, and crosswalks -- yielding a high-resolution, topologically meaningful HD map suitable for downstream driving tasks.
   }
   \label{fig:vm_arc}
\end{figure*}

\begin{figure*}[t]
  \centering
  \includegraphics[width=1\linewidth]{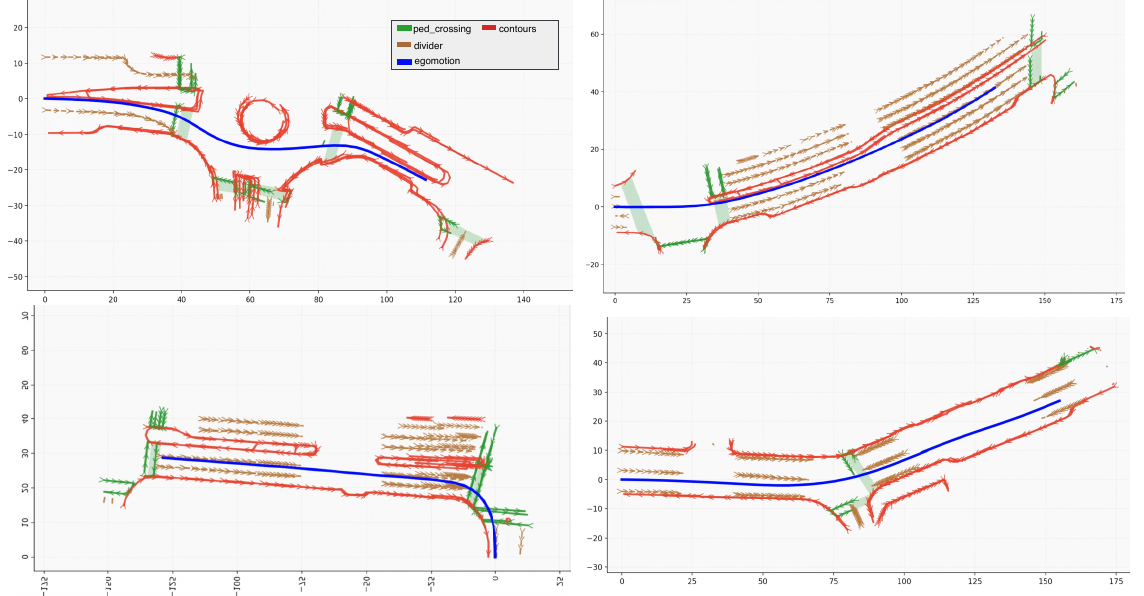}
   \vspace{-0.3cm}
   \caption{Qualitative results of Global HDMap reconstruction results produced by  \textbf{OpenVO} + modified monocular VectorMapNet \cite{liu2023vectormapnet}. Local mapping outputs are gradually fused through OpenVO’s ego-to-world pose estimates, producing a coherent global HD-map reconstruction of the full scenario. We would like to refer to \textcolor{OliveGreen}{\textbf{our supplementary videos}} for further details of the OpenVO-enabled monocular-based global map reconstruction.} 
   \label{fig:quali_hdmap_global}
\end{figure*}

\begin{figure*}[t]
\vspace*{-1em}
  \centering
  \includegraphics[width=1\linewidth]{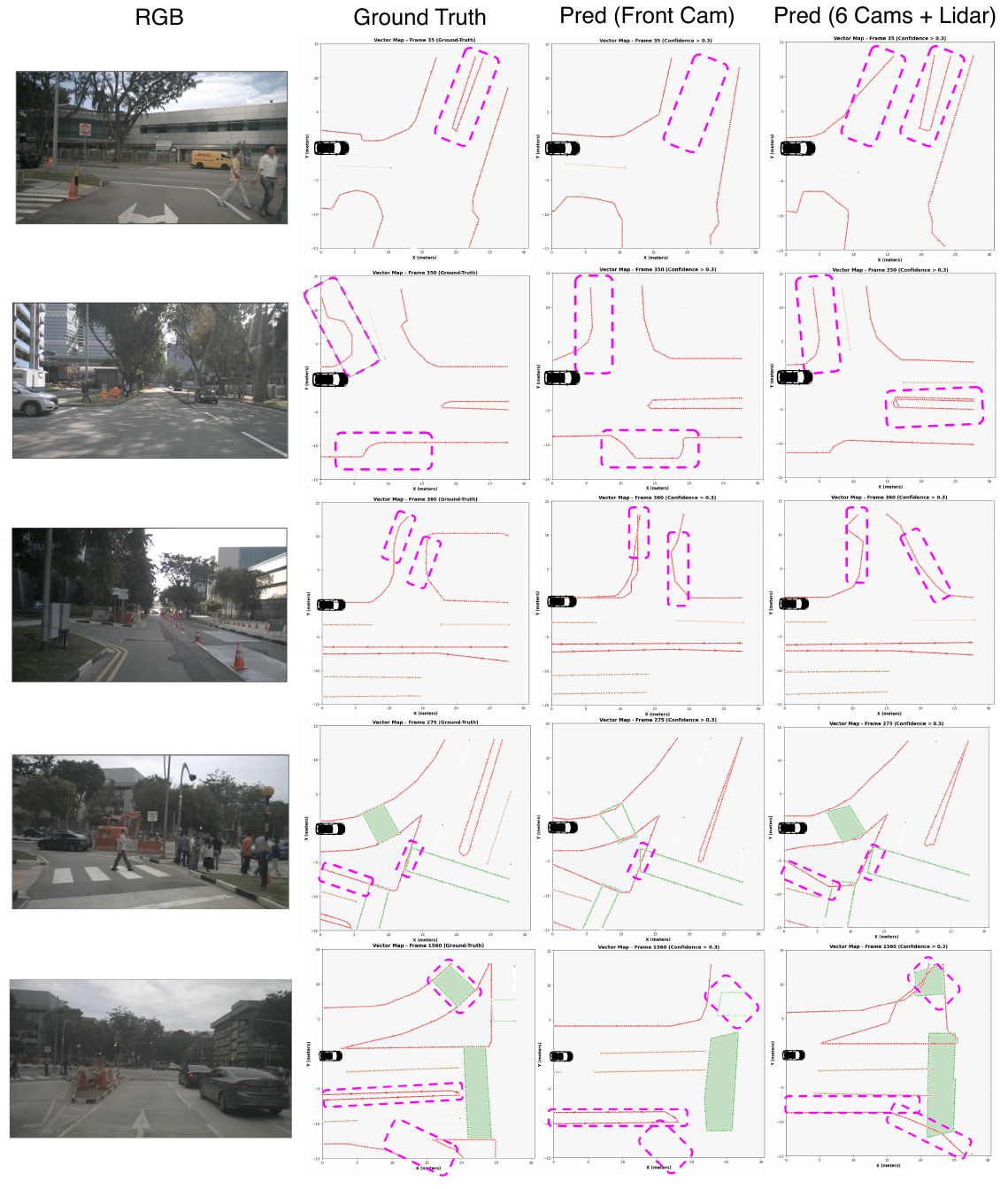}
   \vspace{-0.3cm}
   \caption{Qualitative HD Map reconstruction results produced by modified monocular VectorMapNet \cite{liu2023vectormapnet}. The leftmost column displays the input RGB frames; the second column shows the ground-truth HDMaps; the third column presents the results from our modified VectorMapNet using only a single front-camera image; and the last column shows the outputs under the six-camera + LiDAR configuration. We highlight the differences in each example using \textcolor{magenta}{dashed pink regions}.
   }
   \label{fig:quali_hdmap}
\end{figure*}

Following VectorMapNet~\cite{liu2023vectormapnet}, we reconstruct vectorized local HD maps directly from front-view imagery in ~\cref{fig:vm_arc}. We disregard the LiDAR branch and train the network using only a monocular front-view stream. Ground-truth camera intrinsics and poses are used during training, whereas at inference we rely on the VO estimated by OpenVO and the camera parameters predicted by WildCamera~\cite{zhu2023wildcamera}. For reference purpose, we present our quantitative results on local HDMap reconstruction in~\cref{tab:hdmap_front} and qualitative result in~\cref{fig:quali_hdmap}. Since the network predicts local map fragments independently, we apply post-processing heuristics to spatially align and merge these fragments into a coherent global map. Combining the modified architecture with our proposed OpenVO, we can obtain
the final global map as illustrated in ~\cref{fig:quali_hdmap_global}

\begin{figure*}[t]
\vspace*{-1em}
  \centering
  \includegraphics[width=\linewidth]{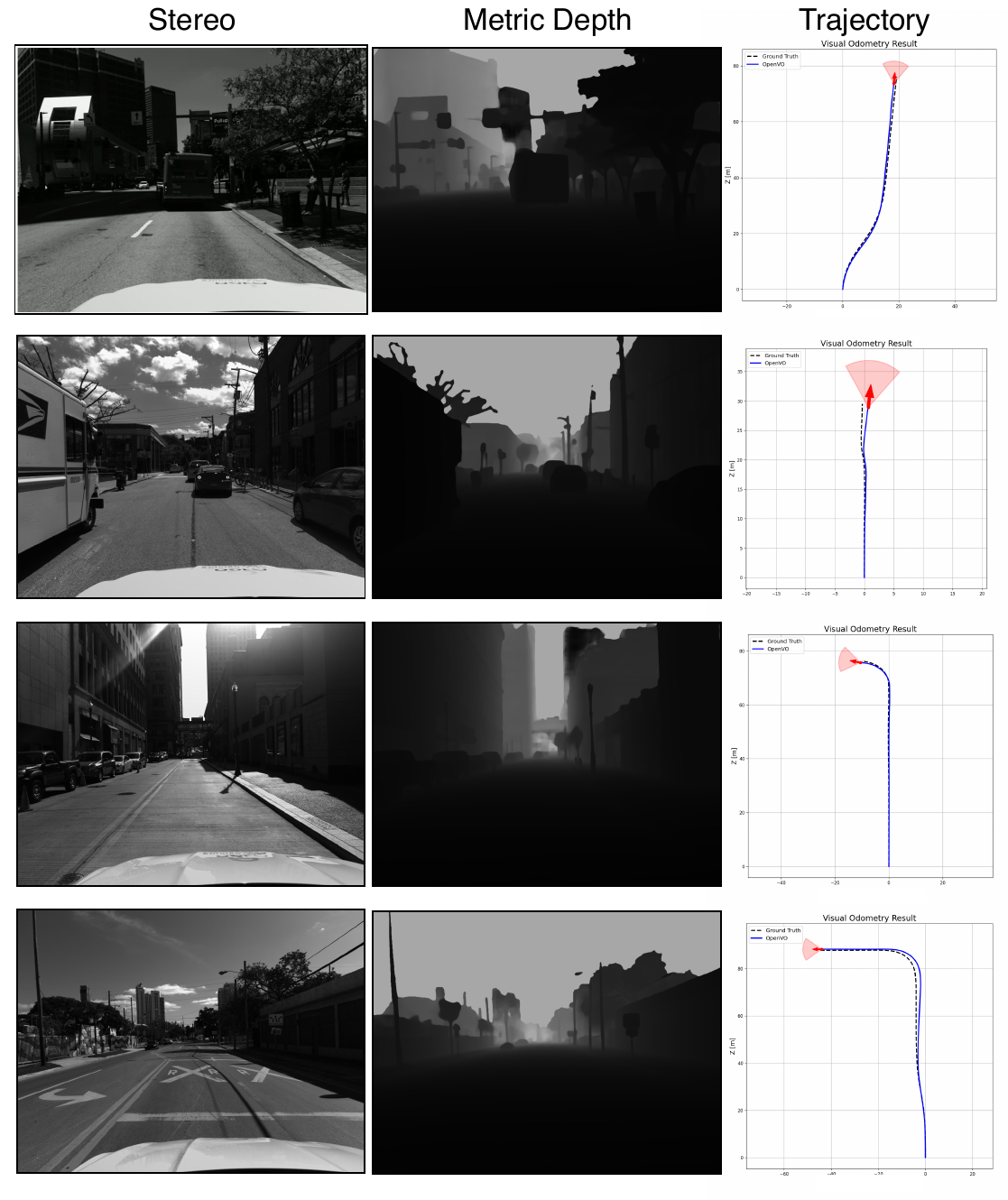}
   \vspace{-0.3cm}
   \caption{\textbf{Qualitative of Stereo benchmark on Argoverse 2 \cite{wilson2021argoverse}.} Each row shows one example, including the input stereo image and the reference metric depth. The stereo images in Argoverse 2 often provides low-quality or weakly constrained metric depth due to limited disparity in long-range regions and visually challenging street scenes. This degradation leads to information loss and introduces uncertainty into downstream VO estimation.
   }
   \label{fig:sup_quali_av}
   \vspace{-0.5cm}
\end{figure*}

\begin{figure*}[t]
\vspace*{-1em}
  \centering
  \includegraphics[width=0.9\linewidth]{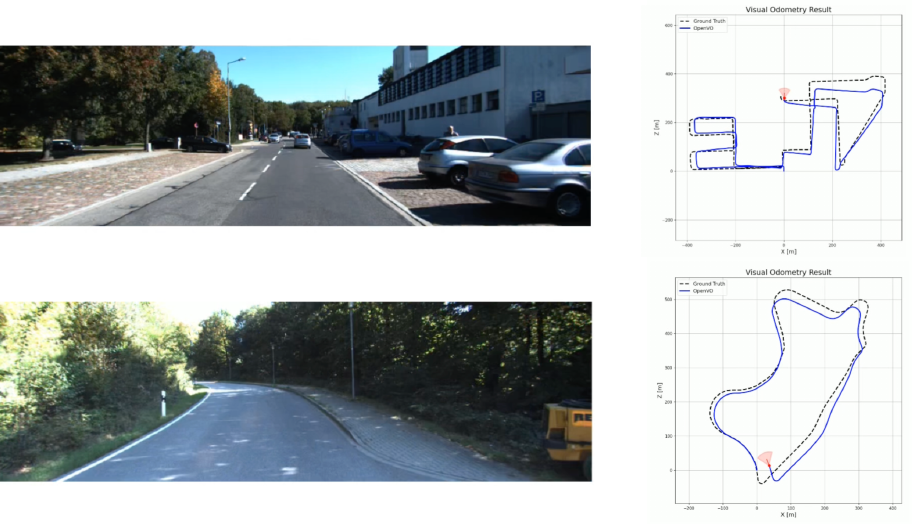}
   \vspace{-0.3cm}
   \caption{\textbf{Qualitative results on KITTI \cite{geiger2013kitti} benchmark.} Each row presents one example. The KITTI camera provides a wider field of view than most datasets, allowing it to capture a richer set of dynamic objects while still preserving its long-range odometry characteristics.}
   \label{fig:sup_quali_kitti}
\end{figure*}

\begin{figure*}[t]
  \centering
  \includegraphics[width=0.85\linewidth]{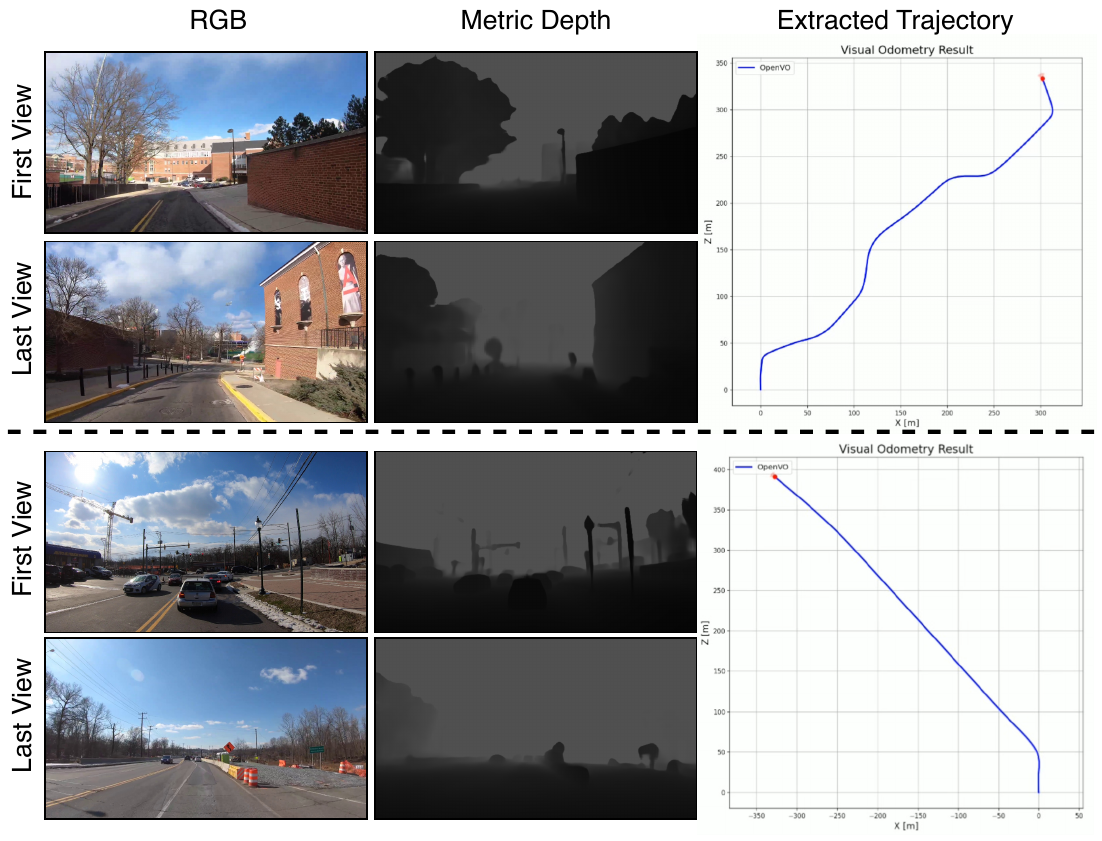}
   \vspace{-0.3cm}
   \caption{\textbf{Qualitative results on real-world captured videos.} We present two examples, each accompanied by the corresponding RGB frames and reference metric-depth images. Real-world videos commonly exhibit numerous environmental artifacts—such as noise, clutter, and dynamic elements—which pose significant challenges for generalizability and real-world performance assessment.
   }
   \label{fig:sup_quali_us}
   \vspace{-0.5cm}
\end{figure*}

\begin{table}[t]
\centering
\setlength\tabcolsep{1.5pt}
\renewcommand{\arraystretch}{1.15}
\begin{tabular}{l|c|>{\columncolor{green!15}}c}
\toprule
\textbf{Metrics} & 
\textbf{Full (6) cams + Lidar} & 
\textbf{Front cam only} \\
\midrule
mAP @ 0.5 ($\uparrow$) & 0.187 & 0.102 \\
mAP @ 1.0 ($\uparrow$) & 0.479 & 0.318 \\
mAP @ 1.5 ($\uparrow$) & 0.646 & 0.472 \\
\textbf{Overall mAP ($\uparrow$)} & \textbf{0.4375} & \textbf{0.297} \\
\midrule
mAP ped @ 1.5 ($\uparrow$) & 0.644 & 0.469 \\
mAP div @ 1.5 ($\uparrow$) & 0.642 & 0.509 \\
mAP cont @ 1.5 ($\uparrow$) & 0.653 & 0.437 \\
mAP @ 1.5 ($\uparrow$) & \textbf{0.646} & \textbf{0.472} \\
\bottomrule
\end{tabular}

\caption{Comparison between the full-sensor configuration and our monocular front-camera–only configuration for HDMap reconstruction. \colorbox{green!15}{Green cells} highlight our configuration. \textit{ped} denotes pedestrian crossings, \textit{div} denotes dividers, and \textit{cont} denotes contours. Performance is reported in terms of mean Average Precision (mAP).} 
\label{tab:hdmap_front}
\vspace{-10pt}
\end{table}

\section{Qualitative Results}
\label{sec:qualiresults}


\myheading{Stereo VO: }We showcase the performance of our OpenVO on the challenging stereo benchmark Argoverse 2 \cite{wilson2021argoverse} in~\cref{fig:sup_quali_av}. In Argoverse 2, the stereo image pairs produce loose and noisy metric depth due to limited baseline, challenging lighting, and frequent low-texture or distant regions. This makes the recovered depth unstable across frames, which in turn causes conventional VO systems to become highly sensitive to depth noise and scale fluctuations. As a result, even small stereo-depth errors can propagate and lead to inconsistent trajectory estimates. In contrast, our OpenVO approach remains robust under these conditions and delivers stable, consistent results despite the imperfections of the stereo-derived metric depth.

\myheading{Long-Range VO: }We show the performance of our OpenVO on the challenging long-range KITTI \cite{geiger2013kitti} benchmark in~\cref{fig:sup_quali_kitti}. The KITTI odometry dataset contains many long-range highway and suburban scenes where most structures lie far from the camera. In such settings, monocular cues become weak: distant objects provide very small pixel motion, depth becomes highly ambiguous, and small errors in these regions can accumulate into noticeable scale drift. As a result, conventional monocular VO systems tend to be sensitive on KITTI, especially over long trajectories where scale inconsistency quickly compounds. Despite these challenges, OpenVO produces stable and consistent results on KITTI, showing that our time-aware and geometry-aware design remains reliable even in long-range, low-parallax environments.

\myheading{Real-World VO: } We further assess the generalizability of OpenVO using videos from real-world environments, as illustrated in~\cref{fig:sup_quali_us}. Real-world driving scenes are heavily populated with vehicles, pedestrians, and cyclists, and are often dominated by occlusions and rapid appearance changes. Such dynamic factors degrade monocular geometric cues and introduce unstable depth signals, making conventional VO pipelines prone to drift and inconsistency. Despite such challenges, OpenVO produces stable and coherent trajectories, highlighting its robustness to noise, clutter, and complex dynamic activity commonly encountered in real-world driving.

\myheading{Application beyond VO: OpenVO-enabled Monocular Global Map Reconstruction: } Reconstructing real-world long-tail scenarios is crucial for safety-critical domains such as autonomous driving, where realistic simulations of rare events provide insights into failure modes, vehicle dynamics, and hazardous scene geometry. However, collecting such data directly is extremely challenging due to economic constraints, safety risks, liability concerns, and legal prohibitions. In contrast, dashcam videos offer an abundant source of real-world long-tail footage, but are typically monocular, uncalibrated, and can be captured at different frame rates, which together pose great challenges for accurate 3D reconstruction tasks like mapping. Prior online mapping approaches~\cite{li2022hdmapnet, liu2023vectormapnet, yuan2024streammapnet} address local mapping using calibrated multi-camera or LiDAR setups with fixed observation rates, which effectively enhance online situational awareness but cannot reconstruct full HD map trajectories or scene evolution required for simulating rare events. In~\cref{fig:quali_hdmap} and ~\cref{fig:quali_hdmap_global} and  and our corresponding supplementary video, we show that \textbf{OpenVO can bridge the global mapping gap} by generating accurate world-consistent poses and fusing them with local mapping to reconstruct the full observable scene, yielding coherent geometric and dynamic information about rare events. This facilitates scalable, monocular-based reconstruction of complex real-world scenarios from dashcam footage, offering deeper insights and enabling more comprehensive training and validation of autonomous driving algorithms in long-tail settings.

